\documentclass[11pt]{article}
\newcommand{\sd}[1]{\mathrel{\scriptstyle \pm #1}} 

\usepackage[final]{acl}

\usepackage{times}
\usepackage{latexsym}

\usepackage[T1]{fontenc}

\usepackage[utf8]{inputenc}

\usepackage{microtype}

\usepackage{inconsolata}
\usepackage[utf8]{inputenc} 
\usepackage[T1]{fontenc}    
\usepackage{hyperref}       
\usepackage{url}            
\usepackage{booktabs}       
\usepackage{amsfonts}       
\usepackage{nicefrac}       
\usepackage{microtype}      
\usepackage{xcolor}         
\usepackage{tcolorbox}
\usepackage{mdframed}
\usepackage{caption}
\usepackage{tabularx} 
\usepackage{colortbl}
\usepackage{xcolor}
\usepackage{collcell} 
\usepackage{svg}
\usepackage{graphicx}
\usepackage{amsmath}
\usepackage{subcaption}
\usepackage{etoc}
\newcommand{\ApplyGradient}[1]{%
    \ifdim #1 pt > 0.8pt \cellcolor{green!40}{#1}%
    \else\ifdim #1 pt > 0.5pt \cellcolor{green!15}{#1}%
    \else\ifdim #1 pt > 0.2pt \cellcolor{orange!20}{#1}%
    \else \cellcolor{red!20}{#1}%
    \fi\fi\fi
}

\newcommand\blfootnote[1]{%
  \begingroup
  \renewcommand\thefootnote{}\footnote{#1}%
  \addtocounter{footnote}{-1}%
  \endgroup
}

\title{Funny or Persuasive, but Not Both: Evaluating Fine-Grained Multi-Concept Control in LLMs}

\newcommand{\inst}[1]{\textsuperscript{#1}}

\author{
  \textbf{Arya Labroo}\inst{1} \:\:
  \textbf{Ivaxi Sheth}\inst{2} \:\:
  \textbf{Vyas Raina}\inst{3} \:\:
  \textbf{Amaani Ahmed}\inst{4} \:\:
  \textbf{Mario Fritz}\inst{2} \:\:
  \\[0.5em]
\inst{1}University of Cambridge \quad
\inst{2}CISPA Helmholtz Center for Information Security \\
\inst{3}Apta \quad
\inst{4}Royal Holloway, University of London
\\[0.5em]
\texttt{al2135@cam.ac.uk} \quad
\texttt{ivaxi.sheth@cispa.de} \quad
\texttt{vyas@apta.chat}
}

\begin{document}
\maketitle

\begin{abstract}
Large Language Models (LLMs) offer strong generative capabilities, but many applications require explicit and \textit{fine-grained} control over specific textual concepts, such as humor, persuasiveness, or formality. Prior approaches in prompting and representation engineering can provide coarse or single-attribute control, but systematic evaluation of multi-attribute settings remains limited. We introduce an evaluation framework for fine-grained controllability for both single- and dual-concept scenarios, focusing on linguistically distinct concept pairs (e.g., persuasiveness vs.~humor). Surprisingly, across multiple LLMs and generative tasks, we find that performance often drops in the dual-concept setting, even though the chosen concepts should in principle be separable. This reveals a fundamental limitation of naive prompting-based control: models struggle with compositionality even when concepts are intuitively independent. Our framework provides systematic evidence of this gap and offers a principled approach for measuring the ability of future methods for multi-concept control.\blfootnote{\url{https://github.com/pencilcase42/finegrained-multiconcept-control}}
\end{abstract}

\section{Introduction}

\begin{figure}[t!]
    \centering
    \includegraphics[width=\linewidth]{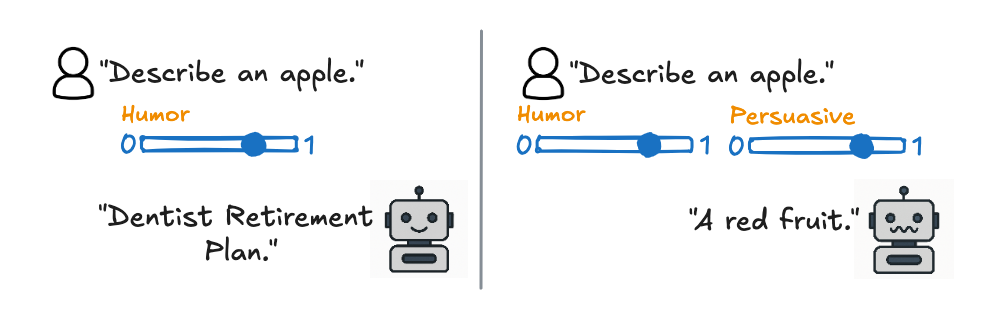} 
    \caption{Illustrative example: an LLM can perform single-concept control, but the explicit presence of a second concept at the input can compromise the ability of the model to control the former concept in its response.}
    \label{fig:teaser}
\end{figure}

Large Language Models (LLMs) are increasingly used in applications such as chat assistants, creative writing, education, and decision support~\cite{achiam2023gpt, brooks2024video, jia2024decision,singhal2025toward, lee2024survey, team2024learnlm,bashiri2024transformative}. Beyond standard text generation, users often desire outputs that exhibit specific styles or concepts~\cite{sun2023evaluating}. For example, a user may wish to rephrase an email to sound more persuasive or funny. More importantly, users often prefer \emph{fine-grained control} over the degree to which such stylistic \emph{concepts}, like humor or persuasiveness, appear in the generated text~\cite{nguyen2025multi, zhang2025controlling}. Furthermore, users may want to modulate multiple concepts. For example, a user may want to increase the humor slightly while maintaining a moderate level of persuasiveness (Figure \ref{fig:teaser}). 

Prior work has explored control through prompting and decoding guided~\cite{brown2020language,pplm,gedi,fudge,yang2023tailor}, representation engineering~\cite{zou2023representation,caa}, and style transfer~\cite{shen2017style,prabhumoye2018style}. These methods demonstrate coarse or single-attribute control, and in some cases enable smooth calibration along one dimension (e.g., SteerLM~\cite{steerlm}, CAA~\cite{caa}). However, systematic and explicit evaluation of multi-concept fine-grained control remains unexplored. Existing benchmarks such as SCTG~\cite{sctg} assess calibration for one attribute at a time, but do not consider how models behave when two distinct concepts are controlled simultaneously.

To address the lack of dual-concept evaluation, we introduce a systematic framework for assessing fine-grained controllability in both single- and dual-concept settings. We study six linguistically distinct concepts—humor, persuasiveness, clarity, politeness, assertiveness, and formality—and deliberately pair concepts that should, in principle, be independent (e.g., clarity vs.~humor). Our experiments use medium-sized instruction-tuned models (7B--14B), prompted across five discrete levels (0--4), with outputs judged via pairwise comparisons by a stronger LLM. Rank correlations between intended and judged levels provide a robust measure of controllability across single- and dual-concept conditions.

In this work, we opt to evaluate prompting, both due to it being a widely accessible method of control, as well have been shown to perform better than many more complicated representation engineering methods proposed in literature for single-concept control~\cite{wu2025axbench}. Our findings are insightful: while prompting achieves sensible fine-grained calibration for individual concepts, performance often \textbf{drops sharply in the dual-concept setting}, even for pairs that are intuitively orthogonal. This suggests that concept dimensions are entangled in ways that resist naive composition. 

More broadly, our evaluation framework is model- and method-agnostic, providing a standardized way to measure controllability across future techniques. By establishing clear metrics and identifying common failure modes, we aim to encourage the development of more robust methods that enable interpretable, multidimensional stylistic control in language models.

\section{Fine-grained Control Evaluation Framework}\label{sec:method}

We define the task of fine-grained concept control as follows. Let $\mathcal{C}$ denote the set of controllable concepts, where each $C \in \mathcal{C}$ represents a semantic dimension such as \texttt{humor} or \texttt{formality}. Each concept $C$ is associated with a discrete scale of levels $\mathcal{L} = \{0,1,\dots,L\}$, where $\ell=0$ denotes no presence and $\ell=L$ denotes maximal presence of the concept. The objective is to evaluate the fine-grained control abilities of a language generation model, $\mathcal G(\cdot)$.

\paragraph{Single-concept control.} Given a textual context $x$ and a target concept $C_a \in \mathcal{C}$ with desired level $\ell \in \mathcal{L}$, the generation model $G$, produces an output,
\begin{equation}
    y_{\ell} = G(x, C_a, \ell).
\end{equation}
Across all levels $\ell \in \{0,\dots,L\}$, this yields a set of outputs $\{y_0, \dots, y_L\}$. For a perfect model $\mathcal{G}$, the ranking of generations by their realized strength of concept $C_a$ would be strictly monotonic in $\ell$, i.e.\ aligned with the intended order $(0,1,\dots,L)$.

\paragraph{Dual-concept control.} Now consider two concepts $C_a, C_b \in \mathcal{C}$, assumed to be semantically distinct. The user specifies desired levels $(\ell_a, \ell_b) \in \mathcal{L}^2$, and the model generates,
\begin{equation}
    y = G(x, C_a, \ell_a, C_b, \ell_b).
\end{equation}
To assess controllability of $C_a$ while holding $C_b$ fixed at $\ell_b=j$, we obtain generations $\{y_{\ell_a,j}\}_{\ell_a=0}^L$ and measure how well their ranking aligns with the intended order $(0,1,\dots,L)$ for $C_a$. This process is repeated for each $j \in \mathcal{L}$, and the overall performance can be averaged over all fixed levels, $j$, giving a controllability profile of $C_a$ given $C_b$. Evaluation is performed symmetrically with $C_b$ as the target concept. In addition to the fixed-level setting, we also consider a \emph{randomized secondary concept} variant. Here, for each target concept $C_a$, we sample $\ell_b \sim \text{Uniform}(\mathcal{L})$ independently for each generation. This variant tests whether control over $C_a$ is disentangled from the level of $C_b$.  

\paragraph{Judge-based evaluation.} To assess whether the generated outputs $\{y_\ell\}$ follow the intended order, we use a judge model $J$ that performs pairwise comparisons between generations\footnote{In preliminary experiments, we also evaluated a \emph{listwise} single-inference approach with the judge-LLM that ranks all responses in a single inference (with responses presented in randomized order). We observed substantial position bias, where the first-presented sample was disproportionately ranked lowest (Appendix~\ref{app:pairvslist}, Table~\ref{tab:model-preference-fractions}).}. Each pair $(y_i, y_j)$ is presented in both orders to avoid position bias, and we define the preference score as,
\begin{equation}
    s(i,j) \;=\; \tfrac{1}{2}\Big(J(y_i,y_j) + \big(1 - J(y_j,y_i)\big)\Big),
\end{equation}
where $J(y_i,y_j) \in \{0, 0.5, 1\}$ denotes whether the judge considers $y_i$ to exhibit more of the target concept than $y_j$ (with $0.5$ for a tie). By summing the pairwise scores for each $y_\ell$ against other levels, we derive an empirical ranking $\hat{r}$ over $\{y_\ell\}$ and measure correlation with the intended ranking $r = (0,1,\dots,L)$ using Spearman~\cite{Spearman1904} $\rho$ correlation. The overall ability of a generation model $\mathcal{G}(\cdot)$ to perform fine-grained control of the selected concepts is quantified as the average of the correlation metrics across a dataset of $N$ contexts $\{x^{(1)},\dots,x^{(N)}\}$. Letting $\rho^{(n)}$ denote the Spearman correlation for instance $x^{(n)}$, we get $\bar{\rho}=\frac{1}{N}\sum_{n=1}^N \rho^{(n)}$. This aggregated scores summarizes the model’s controllability across the dataset. In all experiments, we set $L=4$, corresponding to five levels of control for each concept.

\paragraph{Aggregation and statistical testing.}
For ease of interpretation, we summarize performance with the mean Spearman correlation $\bar{\rho}=\frac{1}{N}\sum_{n=1}^N \rho^{(n)}$.
However, since correlation coefficients are bounded and nonlinearly scaled, we also compute Fisher $z$-transformed correlations~\citep{fisher1915frequency},
\begin{equation}
z^{(n)} \;=\; \frac{1}{2}\ln\!\left(\frac{1+\rho^{(n)}}{1-\rho^{(n)}}\right),
\end{equation}
and aggregate via $\bar{z}=\frac{1}{N}\sum_{n=1}^N z^{(n)}$.
Appendix~\ref{app:fisherresults} considers the Fisher-transformed aggregates, and Appendix~\ref{app:statresults} considers paired $t$-tests conducted on $\{z^{(n)}\}_{n=1}^N$ when comparing conditions. \newline

In this work, we apply the framework to prompting as an initial but also widely used and effective~\cite{wu2025axbench} control method. However, the evaluation protocol is general and can be applied to bespoke approaches designed for fine-grained or multi-concept control.

\begin{figure*}[t]
  \centering
  \begin{subfigure}[t]{0.24\linewidth}
    \centering
    \includegraphics[width=\linewidth]{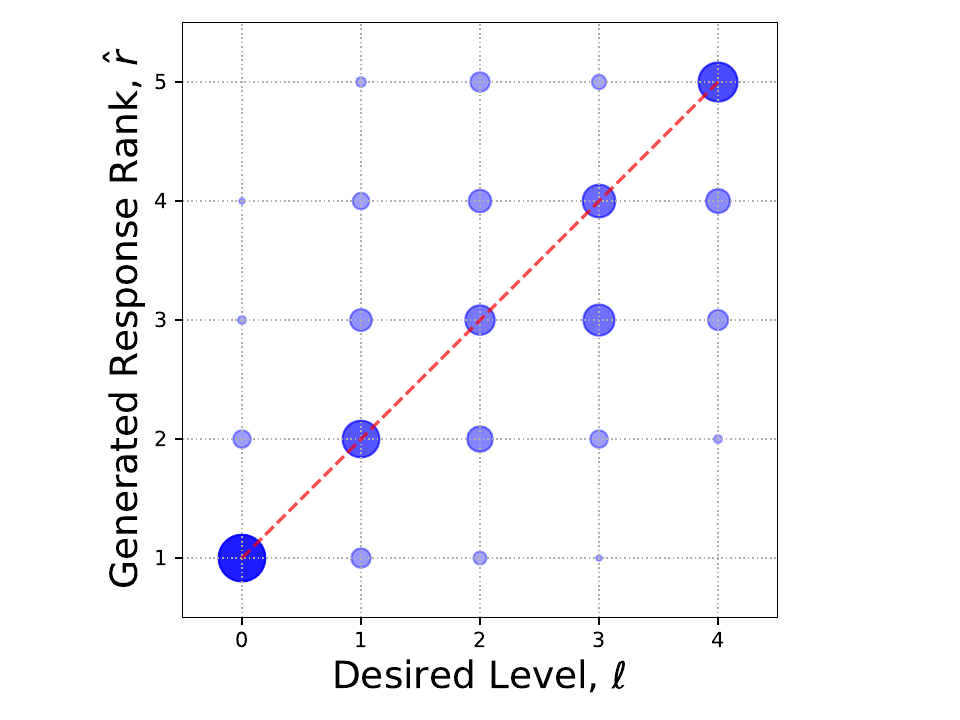}
    \caption{Humor}
  \end{subfigure}\hfill
  \begin{subfigure}[t]{0.24\linewidth}
    \centering
    \includegraphics[width=\linewidth]{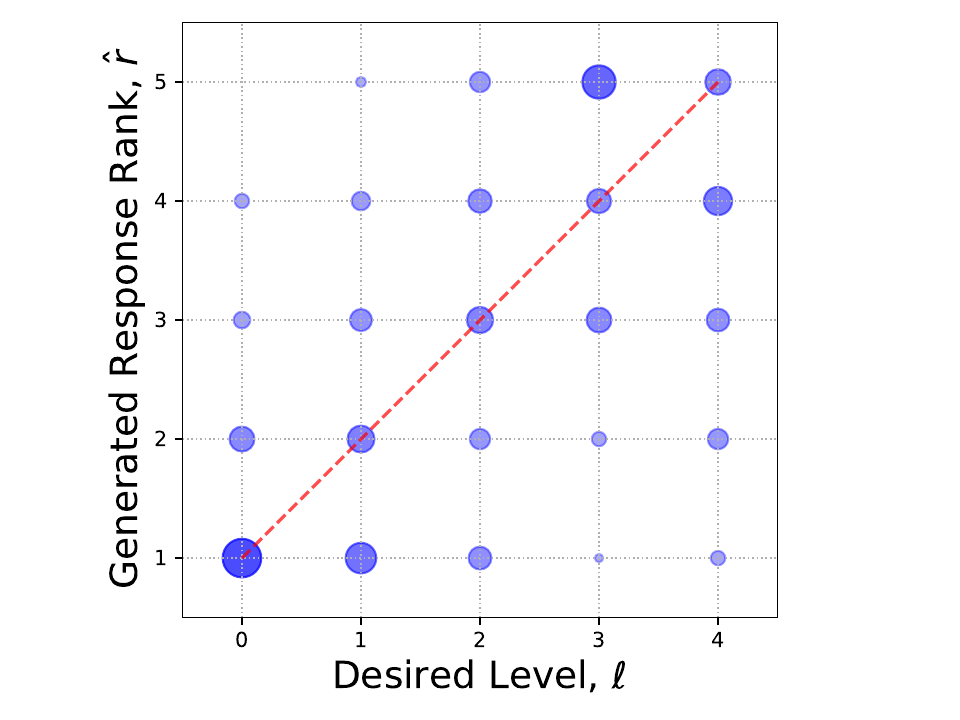}
    \caption{Humor | Persuasiveness}
  \end{subfigure}\hfill
  \begin{subfigure}[t]{0.24\linewidth}
    \centering
    \includegraphics[width=\linewidth]{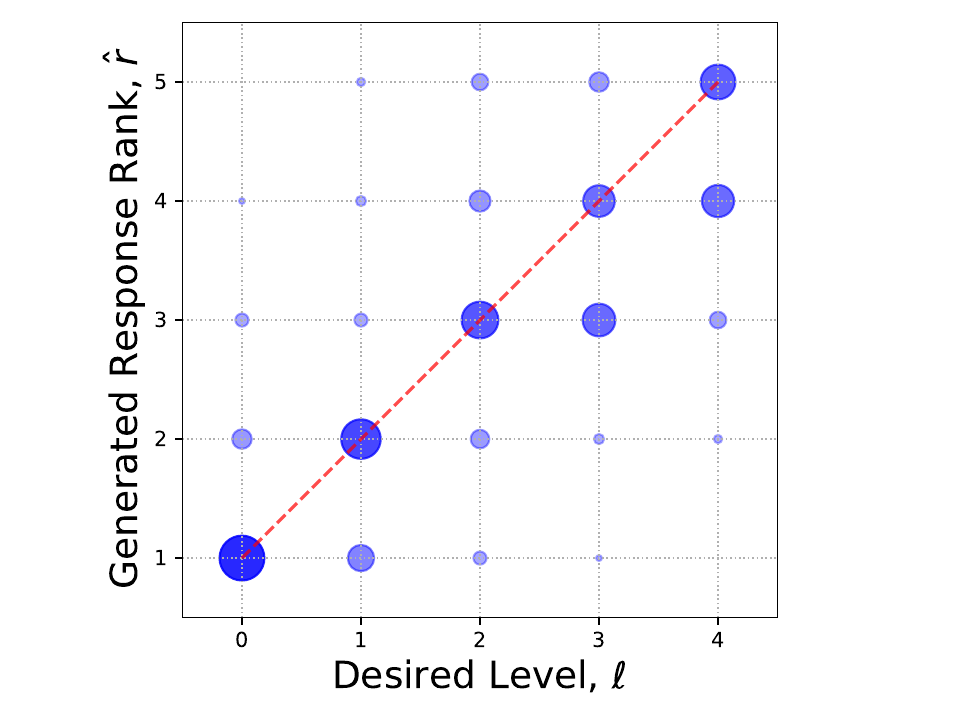}
    \caption{Persuasiveness}
  \end{subfigure}\hfill
  \begin{subfigure}[t]{0.24\linewidth}
    \centering
    \includegraphics[width=\linewidth]{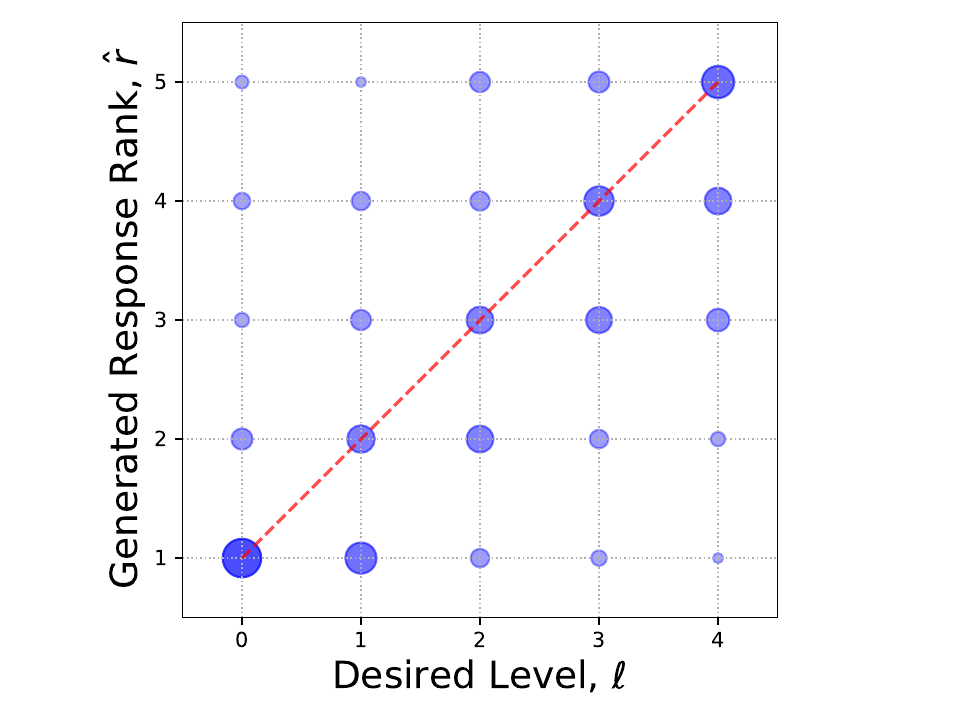}
    \caption{Persuasiveness | Humor}
  \end{subfigure}
  \caption{Model-generated response rank of the target concept versus the desired level. Point size and density indicate the number of samples at each coordinate. Results shown for Llama-11B with the secondary concept level \textit{randomly} sampled. For example, ``Humor $|$ Persuasiveness'' denotes responses generated independently for each humor level (target concept) while persuasiveness is randomly set for each inference.}
  \label{fig:main-scatter}
\end{figure*}

\section{Experiments}

\subsection{Setup}
\paragraph{Models.}  We evaluate medium-sized, instruction-tuned LLMs in the 10B--14B parameter range: Llama 3.2-11B~\cite{meta2024llama}, Gemma 3-12B~\cite{team2025gemma}, and Qwen3-14B~\cite{qwen3}. These models are representative of widely deployed generation systems that are computationally affordable while still capable of complex stylistic control. We used GPT-4.1~\cite{openai2023gpt} as the judge-LLM. To validate the judge, we performed human validation, where we observed that the judge was fairly aligned with the human participants (see Appendix \ref{app:humaneval}).
In Appendix~\ref{app:allresults}, we extend our evaluation to smaller models.


\paragraph{Data and Concepts.} 
We consider three tasks: argument generation, story generation, and structured text generation, each with 75 unique test samples. For argument generation, we use the Persuasion dataset~\cite{durmus2024persuasion}. We discard the associated arguments and scores, using each claim as a prompt for generating an \textit{argument} controlled across different stylistic and pragmatic dimensions. For story generation, we use the ROCStories dataset~\cite{mostafazadeh2016corpus}, each example begins with the same narrative prompt, and the model continues the story in the requested styles. For structured text generation, we provide structured inputs from the GEM dataset~\cite{gehrmann2021gem} that must be converted into textual descriptions, testing the model’s ability to verbalize and stylistically adapt structured information. 

We evaluate six concepts: humor, persuasiveness, clarity, politeness, assertiveness, and formality. These were selected for their (i) relevance to real-world applications, (ii) linguistic distinctiveness supported by factor-analytic studies~\cite{nevid1979factor,kearney1984factor}, and (iii) practical motivation for independent adjustment (e.g., writing assistants, educational tools, debate preparation). For multi-concept evaluation, we study three pairs: humor--persuasiveness, clarity--politeness, and assertiveness--formality, chosen because theoretical and empirical evidence suggests they are distinct dimensions~\cite{biber1995dimensions, baror2022sociolinguisticsbasedframeworkstudypoliteness}. 

Importantly, our evaluation does not require these concepts to be disentangled in a model's internal representation. The only assumption is user-facing: the concepts are sufficiently distinguishable to annotators and end users to support separate specification (e.g., ``high clarity, low politeness''). Whether a model internally entangles these dimensions is orthogonal to this requirement. Accordingly, our conclusions do not assume conceptual separability: even under strong internal entanglement, an effective control method should still track user-specified levels for each concept without substantial cross-concept interference.

To achieve fine-grained control over single and dual-concept levels, we design structured prompt templates that explicitly encode the desired concept intensities; detailed templates and examples are provided in Appendix~\ref{app:prompts}.

\begin{table*}[htb!]
\centering
\scriptsize
\setlength{\tabcolsep}{4pt} 
\begin{tabular}{l*{3}{ccc}*{3}{ccc}*{3}{ccc}}
\toprule
\multicolumn{1}{c}{} &
\multicolumn{3}{c}{\textbf{Argument Generation}} &
\multicolumn{3}{c}{\textbf{Story Generation}} &
\multicolumn{3}{c}{\textbf{Structured Text Generation}} \\
\cmidrule(lr){2-4}\cmidrule(lr){5-7}\cmidrule(lr){8-10}
& \textbf{Llama-11B} & \textbf{Gemma-12B} & \textbf{Qwen-14B}
& \textbf{Llama-11B} & \textbf{Gemma-12B} & \textbf{Qwen-14B}
& \textbf{Llama-11B} & \textbf{Gemma-12B} & \textbf{Qwen-14B} \\
\midrule
$C_a$ (single)          & $0.76{\sd{0.23}}$ & $0.95{\sd{0.07}}$ & $0.92{\sd{0.11}}$
                        & $0.81{\sd{0.26}}$ & $0.95{\sd{0.06}}$ & $0.92{\sd{0.10}}$
                        & $0.73{\sd{0.22}}$ & $0.94{\sd{0.12}}$ & $0.90{\sd{0.12}}$ \\
$C_a \mid C_b$ fixed    & $0.51{\sd{0.41}}$ & $0.88{\sd{0.14}}$ & $0.88{\sd{0.15}}$
                        & $0.36{\sd{0.45}}$ & $0.81{\sd{0.22}}$ & $0.90{\sd{0.12}}$
                        & $0.31{\sd{0.50}}$ & $0.88{\sd{0.15}}$ & $0.84{\sd{0.19}}$ \\
$C_a \mid C_b$ rand     & $0.54{\sd{0.35}}$ & $0.83{\sd{0.20}}$ & $0.88{\sd{0.16}}$
                        & $0.33{\sd{0.49}}$ & $0.74{\sd{0.25}}$ & $0.88{\sd{0.15}}$
                        & $0.17{\sd{0.48}}$ & $0.79{\sd{0.21}}$ & $0.81{\sd{0.21}}$ \\
\midrule
$C_b$ (single)          & $0.81{\sd{0.22}}$ & $0.98{\sd{0.04}}$ & $0.96{\sd{0.05}}$
                        & $0.80{\sd{0.19}}$ & $0.97{\sd{0.04}}$ & $0.93{\sd{0.10}}$
                        & $0.89{\sd{0.14}}$ & $0.99{\sd{0.02}}$ & $0.99{\sd{0.03}}$ \\
$C_b \mid C_a$ fixed    & $0.58{\sd{0.38}}$ & $0.83{\sd{0.19}}$ & $0.84{\sd{0.18}}$
                        & $0.59{\sd{0.35}}$ & $0.69{\sd{0.34}}$ & $0.85{\sd{0.18}}$
                        & $0.56{\sd{0.41}}$ & $0.91{\sd{0.15}}$ & $0.90{\sd{0.14}}$ \\
$C_b \mid C_a$ rand     & $0.52{\sd{0.40}}$ & $0.76{\sd{0.21}}$ & $0.81{\sd{0.21}}$
                        & $0.58{\sd{0.34}}$ & $0.70{\sd{0.31}}$ & $0.83{\sd{0.20}}$
                        & $0.51{\sd{0.39}}$ & $0.79{\sd{0.19}}$ & $0.83{\sd{0.19}}$ \\
\bottomrule
\end{tabular}
\caption{\textbf{Humor--persuasiveness.} Spearman correlations for single-concept and dual-concept (fixed / random) across argument, story, and structured text generation.}
\label{tab:humor-pers-large}
\end{table*}

\begin{table*}[htb!]
\centering
\scriptsize
\setlength{\tabcolsep}{4pt} 
\begin{tabular}{l*{3}{ccc}*{3}{ccc}*{3}{ccc}}
\toprule
\multicolumn{1}{c}{} &
\multicolumn{3}{c}{\textbf{Argument Generation}} &
\multicolumn{3}{c}{\textbf{Story Generation}} &
\multicolumn{3}{c}{\textbf{Structured Text Generation}} \\
\cmidrule(lr){2-4}\cmidrule(lr){5-7}\cmidrule(lr){8-10}
& \textbf{Llama-11B} & \textbf{Gemma-12B} & \textbf{Qwen-14B} 
& \textbf{Llama-11B} & \textbf{Gemma-12B} & \textbf{Qwen-14B}
& \textbf{Llama-11B} & \textbf{Gemma-12B} & \textbf{Qwen-14B} \\
\midrule
$C_a$ (single)          & $-0.02{\sd{0.52}}$ & $0.52{\sd{0.46}}$ & $0.65{\sd{0.30}}$
                        & $0.45{\sd{0.46}}$  & $0.92{\sd{0.11}}$ & $0.89{\sd{0.12}}$
                        & $0.21{\sd{0.56}}$  & $0.15{\sd{0.61}}$ & $0.64{\sd{0.21}}$ \\
$C_a \mid C_b$ fixed    & $0.02{\sd{0.53}}$  & $0.02{\sd{0.56}}$ & $0.64{\sd{0.34}}$
                        & $-0.01{\sd{0.53}}$ & $0.35{\sd{0.43}}$ & $0.74{\sd{0.26}}$
                        & $0.02{\sd{0.45}}$  & $-0.25{\sd{0.53}}$ & $0.39{\sd{0.43}}$ \\
$C_a \mid C_b$ rand     & $-0.05{\sd{0.51}}$ & $0.12{\sd{0.50}}$ & $0.63{\sd{0.32}}$
                        & $-0.07{\sd{0.50}}$ & $0.29{\sd{0.47}}$ & $0.64{\sd{0.29}}$
                        & $0.08{\sd{0.40}}$  & $-0.19{\sd{0.45}}$ & $0.38{\sd{0.43}}$ \\
\midrule
$C_b$ (single)          & $0.76{\sd{0.25}}$  & $0.95{\sd{0.07}}$ & $0.93{\sd{0.10}}$
                        & $0.84{\sd{0.21}}$  & $0.98{\sd{0.03}}$ & $0.96{\sd{0.07}}$
                        & $0.73{\sd{0.28}}$  & $0.97{\sd{0.03}}$ & $0.93{\sd{0.09}}$ \\
$C_b \mid C_a$ fixed    & $0.76{\sd{0.25}}$  & $0.83{\sd{0.19}}$ & $0.88{\sd{0.14}}$
                        & $0.71{\sd{0.30}}$  & $0.86{\sd{0.17}}$ & $0.95{\sd{0.08}}$
                        & $0.45{\sd{0.42}}$  & $0.79{\sd{0.31}}$ & $0.79{\sd{0.26}}$ \\
$C_b \mid C_a$ rand     & $0.77{\sd{0.29}}$  & $0.80{\sd{0.18}}$ & $0.84{\sd{0.15}}$
                        & $0.71{\sd{0.31}}$  & $0.72{\sd{0.26}}$ & $0.92{\sd{0.08}}$
                        & $0.37{\sd{0.47}}$  & $0.63{\sd{0.33}}$ & $0.76{\sd{0.28}}$ \\
\bottomrule
\end{tabular}
\caption{\textbf{Clarity--politeness.} Spearman correlations for single-concept and dual-concept (fixed / random) across argument, story, and structured text generation.}
\label{tab:clar-polit-large}
\end{table*}

\begin{table*}[htb!]
\centering
\scriptsize
\setlength{\tabcolsep}{4pt} 
\begin{tabular}{l*{3}{ccc}*{3}{ccc}*{3}{ccc}}
\toprule
\multicolumn{1}{c}{} &
\multicolumn{3}{c}{\textbf{Argument Generation}} &
\multicolumn{3}{c}{\textbf{Story Generation}} &
\multicolumn{3}{c}{\textbf{Structured Text Generation}} \\
\cmidrule(lr){2-4}\cmidrule(lr){5-7}\cmidrule(lr){8-10}
& \textbf{Llama-11B} & \textbf{Gemma-12B} & \textbf{Qwen-14B} 
& \textbf{Llama-11B} & \textbf{Gemma-12B} & \textbf{Qwen-14B}
& \textbf{Llama-11B} & \textbf{Gemma-12B} & \textbf{Qwen-14B} \\
\midrule
$C_a$ (single)          & $0.92{\sd{0.09}}$ & $0.98{\sd{0.03}}$ & $0.99{\sd{0.02}}$
                        & $0.93{\sd{0.09}}$ & $1.00{\sd{0.02}}$ & $0.98{\sd{0.05}}$
                        & $0.80{\sd{0.24}}$ & $0.93{\sd{0.14}}$ & $0.96{\sd{0.07}}$ \\
$C_a \mid C_b$ fixed    & $0.56{\sd{0.40}}$ & $0.97{\sd{0.05}}$ & $0.97{\sd{0.05}}$
                        & $0.77{\sd{0.25}}$ & $0.96{\sd{0.07}}$ & $0.96{\sd{0.06}}$
                        & $0.42{\sd{0.45}}$ & $0.77{\sd{0.33}}$ & $0.88{\sd{0.15}}$ \\
$C_a \mid C_b$ rand     & $0.41{\sd{0.43}}$ & $0.92{\sd{0.10}}$ & $0.94{\sd{0.08}}$
                        & $0.77{\sd{0.23}}$ & $0.96{\sd{0.06}}$ & $0.96{\sd{0.05}}$
                        & $0.22{\sd{0.48}}$ & $0.71{\sd{0.33}}$ & $0.86{\sd{0.17}}$ \\
\midrule
$C_b$ (single)          & $0.75{\sd{0.32}}$ & $0.99{\sd{0.03}}$ & $0.98{\sd{0.03}}$
                        & $0.67{\sd{0.33}}$ & $0.98{\sd{0.04}}$ & $0.97{\sd{0.06}}$
                        & $0.66{\sd{0.32}}$ & $0.95{\sd{0.08}}$ & $0.87{\sd{0.16}}$ \\
$C_b \mid C_a$ fixed    & $0.48{\sd{0.47}}$ & $0.90{\sd{0.12}}$ & $0.94{\sd{0.08}}$
                        & $0.51{\sd{0.42}}$ & $0.93{\sd{0.10}}$ & $0.91{\sd{0.10}}$
                        & $0.43{\sd{0.50}}$ & $0.72{\sd{0.36}}$ & $0.76{\sd{0.29}}$ \\
$C_b \mid C_a$ rand     & $0.45{\sd{0.44}}$ & $0.85{\sd{0.15}}$ & $0.93{\sd{0.07}}$
                        & $0.41{\sd{0.46}}$ & $0.89{\sd{0.12}}$ & $0.89{\sd{0.12}}$
                        & $0.40{\sd{0.51}}$ & $0.72{\sd{0.26}}$ & $0.75{\sd{0.24}}$ \\
\bottomrule
\end{tabular}
\caption{\textbf{Formality--assertiveness.} Spearman correlations for single-concept and dual-concept (fixed / random) across argument, story, and structured text generation.}
\label{tab:form-assert-large}
\end{table*}

\subsection{Results}

Tables~\ref{tab:humor-pers-large}--\ref{tab:form-assert-large} report the average Spearman correlations ($\bar{\rho}$) between intended concept levels and the empirical ranks of generated responses (Section~\ref{sec:method}). Appendix~\ref{app:fisherresults} reports Fisher-transformed aggregates, with paired tests in Appendix~\ref{app:statresults}. For most concept pairs, models generally show strong single-concept control but notable degradation when a secondary concept is introduced. In the humor–persuasiveness pair, this decline is more pronounced in structured text generation. For clarity–politeness, it differs significantly between the tasks. 
For this concept pair, for Llama-11B, argument generation exhibits little control over the clarity concept with near-zero correlation, whereas story generation and structured text generation achieve significantly higher correlations.
Politeness follows the standard pattern: high performance for a single concept, but a drop when clarity is introduced. Similarly, in formality–assertiveness, both concepts exhibit consistently high single-concept control (up to $1.00$ for Gemma) but degrade under dual-control conditions.

\paragraph{General trends.} 
Three broader insights emerge: (i) Qwen-14B and Gemma-12B consistently outperform Llama across all settings. This suggests that larger or more instruction-tuned models better preserve disentanglement between stylistic dimensions. (ii) Dual-concept interference remains a central limitation: even when single-concept control is strong, the introduction of a secondary dimension leads to drops in alignment (Figure~\ref{fig:main-scatter}), suggesting weak compositionality of stylistic control. (iii) Task context strongly modulates controllability. Narrative generation allows more flexible style variation, whereas argumentative and structured contexts amplify conflicts between stylistic goals. Together, these results highlight that current LLMs can vary in style along individual axes but struggle to jointly coordinate multiple stylistic dimensions, despite the styles being theoretically disentangled. Similar trends are observed for structured text generation in Tables~\ref{tab:clar-polit-full}-\ref{tab:humour-pers-full}. Finally, histograms of sample-level correlations (Appendix~\ref{sec:app-hist}) confirm that, with the presence of a second concept, correlations generally decrease across most samples, as opposed to only a few samples skewing the average Spearman correlations ($\bar{\rho}$) reported.

\section{Conclusions}

This work introduced a framework to evaluate fine-grained control of stylistic concepts in LLMs. Through experiments on three pairs of linguistically distinct concepts, we found that while prompting models offers some degree of single-concept controllability. Performance can, however, drop notably in the dual-concept setting even for concept pairs that should, in principle, be disentangled. These findings illustrate that current LLMs struggle to provide fine-grained, disentangled control across multiple stylistic dimensions. We believe this work establishes a foundation for future research on interpretable and compositional concept control. By offering a clear, reproducible benchmark and quantitative metrics, it provides the basis for developing and adapting methodologies for fine-grained multi-concept control.


\section{Limitations}

This study has four main limitations. First, our evaluation focused on three concept pairs (humor--persuasiveness, clarity--politeness, assertiveness--formality). By this, we are examining concept pairs that should, in principle, exhibit no interference. The proposed framework is general and could be applied to a broader range of concept combinations in future work.  

Second, we restricted our analysis to small/medium-sized generation models (3B--14B parameters). These models are widely accessible and computationally practical, but larger LLMs may exhibit different behaviors. Extending the framework to stronger models would provide insight into whether scale improves fine-grained and multi-concept controllability. 

Third, we evaluated only direct prompt-based control. Although prompt-based control is easiest to use in practice and has been shown to be more effective than many representation engineering strategies~\cite{wu2025axbench}, future work could adapt representation-engineering approaches or logit-biasing techniques and then evaluate using the proposed framework in this work, to test their ability to provide precise, multi-level concept control.

\section*{Acknowledgements}
This work was partially funded by ELSA – European Lighthouse on Secure and Safe AI funded by the European Union under grant agreement No. 101070617. This work was also partially supported by the European Union’s Horizon Europe research and innovation programme under grant agreement No 101214398 (ELLIOT). Funded by the European Union. Views and opinions expressed are however those of the author(s) only and do not necessarily reflect those of the European Union or the European Commission. Neither the European Union nor the European Commission can be held responsible for them.

\newpage
\bibliography{ref}

@article{nguyen2025multi,
  title={Multi-attribute steering of language models via targeted intervention},
  author={Nguyen, Duy and Prasad, Archiki and Stengel-Eskin, Elias and Bansal, Mohit},
  journal={arXiv preprint arXiv:2502.12446},
  year={2025}
}

@inproceedings{zhang2025controlling,
  title={Controlling large language models through concept activation vectors},
  author={Zhang, Hanyu and Wang, Xiting and Li, Chengao and Ao, Xiang and He, Qing},
  booktitle={Proceedings of the AAAI Conference on Artificial Intelligence},
  volume={39},
  number={24},
  pages={25851--25859},
  year={2025}
}

@inproceedings{gehrmann2021gem,
  title={The GEM Benchmark: Natural Language Generation, its Evaluation and Metrics},
  author={Gehrmann, Sebastian and Adewumi, Tosin and Aggarwal, Karmanya and Ammanamanchi, Pawan Sasanka and Aremu, Anuoluwapo and Bosselut, Antoine and Chandu, Khyathi Raghavi and Clinciu, Miruna and Das, Dipanjan and Dhole, Kaustubh and others},
  booktitle={Proceedings of the 1st Workshop on Natural Language Generation, Evaluation, and Metrics (GEM 2021)},
  pages={96--120},
  year={2021}
}

@inproceedings{sun2023evaluating,
  title={Evaluating Large Language Models on Controlled Generation Tasks},
  author={Sun, Jiao and Tian, Yufei and Zhou, Wangchunshu and Xu, Nan and Hu, Qian and Gupta, Rahul and Wieting, John and Peng, Nanyun and Ma, Xuezhe},
  booktitle={Proceedings of the 2023 Conference on Empirical Methods in Natural Language Processing},
  pages={3155--3168},
  year={2023}
}

@inproceedings{mostafazadeh2016corpus,
  title={A corpus and cloze evaluation for deeper understanding of commonsense stories},
  author={Mostafazadeh, Nasrin and Chambers, Nathanael and He, Xiaodong and Parikh, Devi and Batra, Dhruv and Vanderwende, Lucy and Kohli, Pushmeet and Allen, James},
  booktitle={Proceedings of the 2016 Conference of the North American Chapter of the Association for Computational Linguistics: Human Language Technologies},
  pages={839--849},
  year={2016}
}

@article{bashiri2024transformative,
  title={Transformative influence of llm and ai tools in student social media engagement: Analyzing personalization, communication efficiency, and collaborative learning},
  author={Bashiri, Masoud and Kowsari, Kamran},
  journal={arXiv preprint arXiv:2407.15012},
  year={2024}
}

@article{team2024learnlm,
  title={LearnLM: Improving Gemini for Learning},
  author={Modi, Abhinit and Veerubhotla, Aditya Srikanth and Rysbek, Aliya and Huber, Andrea and Wiltshire, Brett and Veprek, Brian and Gillick, Daniel and Kasenberg, Daniel and Ahmed, Derek and others},
  journal={arXiv preprint arXiv:2412.16429},
  year={2024}
}

@article{lee2024survey,
  title={A survey of large language models in finance (finllms)},
  author={Lee, Jean and Stevens, Nicholas and Han, Soyeon Caren and Song, Minseok},
  journal={arXiv preprint arXiv:2402.02315},
  year={2024}
}

@article{singhal2025toward,
  title={Toward expert-level medical question answering with large language models},
  author={Singhal, Karan and Tu, Tao and Gottweis, Juraj and Sayres, Rory and Wulczyn, Ellery and Amin, Mohamed and Hou, Le and Clark, Kevin and Pfohl, Stephen R and Cole-Lewis, Heather and others},
  journal={Nature Medicine},
  pages={1--8},
  year={2025},
  publisher={Nature Publishing Group US New York}
}

@article{jia2024decision,
  title={Decision-making behavior evaluation framework for llms under uncertain context},
  author={Jia, Jingru and Yuan, Zehua and Pan, Junhao and McNamara, Paul E and Chen, Deming},
  journal={arXiv preprint arXiv:2406.05972},
  year={2024}
}

@misc{ref1,
  author = {Yi Liu and Xiangyu Liu and Xiangrong Zhu and Wei Hu},
  title = {Multi-Aspect Controllable Text Generation with Disentangled Counterfactual Augmentation},
  journal = {Unknown Journal},
  year = {2024},
  doi = {10.48550/arXiv.2405.19958},
  pages = {9231-9253}
}

@misc{ref2,
  author = {Yuxuan Gu and Xiaocheng Feng and Sicheng Ma and Lingyuan Zhang and Heng Gong and Bing Qin},
  title = {A Distributional Lens for Multi-Aspect Controllable Text Generation},
  journal = {Unknown Journal},
  year = {2022},
  doi = {10.48550/arXiv.2210.02889},
  pages = {1023-1043}
}

@misc{ref4,
  author = {Xuancheng Huang and Zijun Liu and Peng Li and Tao Li and Maosong Sun and Yang Liu},
  title = {An Extensible Plug-and-Play Method for Multi-Aspect Controllable Text Generation},
  journal = {Unknown Journal},
  year = {2022},
  doi = {10.18653/v1/2023.acl-long.849},
  pages = {15233-15256}
}

@misc{ref10,
  author = {Weihao Zeng and Lulu Zhao and Keqing He and Ruotong Geng and Jingang Wang and Wei Wu and Weiran Xu},
  title = {Seen to Unseen: Exploring Compositional Generalization of Multi-Attribute Controllable Dialogue Generation},
  journal = {ArXiv},
  year = {2023},
  doi = {10.48550/arXiv.2306.10317},
  volume = {abs/2306.10317}
}

@misc{ref12,
  author = {Jing Qian and Li Dong and Yelong Shen and Furu Wei and Weizhu Chen},
  title = {Controllable Natural Language Generation with Contrastive Prefixes},
  journal = {Unknown Journal},
  year = {2022},
  doi = {10.18653/v1/2022.findings-acl.229},
  pages = {2912-2924}
}

@misc{ref13,
  author = {Yuxuan Gu and Xiaocheng Feng and Sicheng Ma and Lingyuan Zhang and Heng Gong and Weihong Zhong and Bing Qin},
  title = {Controllable Text Generation via Probability Density Estimation in the Latent Space},
  journal = {ArXiv},
  year = {2022},
  doi = {10.48550/arXiv.2212.08307},
  volume = {abs/2212.08307}
}

@inproceedings{ref14,
  title={Tailor: A soft-prompt-based approach to attribute-based controlled text generation},
  author={Yang, Kexin and Liu, Dayiheng and Lei, Wenqiang and Yang, Baosong and Xue, Mingfeng and Chen, Boxing and Xie, Jun},
  booktitle={Proceedings of the 61st Annual Meeting of the Association for Computational Linguistics (Volume 1: Long Papers)},
  pages={410--427},
  year={2023}
}

@inproceedings{ref20,
  title={DisCup: Discriminator Cooperative Unlikelihood Prompt-tuning for Controllable Text Generation},
  author={Zhang, Hanqing and Song, Dawei},
  booktitle={Proceedings of the 2022 Conference on Empirical Methods in Natural Language Processing},
  pages={3392--3406},
  year={2022}
}

@article{ref31,
  title={Attribute Alignment: Controlling Text Generation from Pre-trained Language Models},
  author={Yu, Dian and Yu, Zhou and Sagae, Kenji},
  journal={Findings of the Association for Computational Linguistics: EMNLP 2021},
  year={2021}
}

@inproceedings{konen2024style,
  title={Style Vectors for Steering Generative Large Language Models},
  author={Konen, Kai and Jentzsch, Sophie and Diallo, Diaoul{\'e} and Sch{\"u}tt, Peer and Bensch, Oliver and El Baff, Roxanne and Opitz, Dominik and Hecking, Tobias},
  booktitle={Findings of the Association for Computational Linguistics: EACL 2024},
  pages={782--802},
  year={2024}
}

@inproceedings{cai2024self,
  title={Self-Control of LLM Behaviors by Compressing Suffix Gradient into Prefix Controller},
  author={Cai, Min and Zhang, Yuchen and Zhang, Shichang and Yin, Fan and Zou, Difan and Yue, Yisong and Hu, Ziniu},
  booktitle={ICML 2024 Workshop on Mechanistic Interpretability}
}

@article{turnersteering,
  title={Steering Language Models with Activation Engineering},
  author={Turner, Alexander Matt and Thiergart, Lisa and Leech, Gavin and Udell, David and Vazquez, Juan J and Mini, Ulisse and MacDiarmid, Monte}
}

@article{article,
author = {Corbett-Davies, Sam and Pierson, Emma and Feller, Avi and Goel, Sharad and Huq, Aziz},
year = {2017},
month = {01},
pages = {},
title = {{Algorithmic decision making and the cost of fairness}}
}

@article{meta2024llama,
  title={Llama 3.2: Revolutionizing edge ai and vision with open, customizable models},
  author={Meta, AI},
  journal={Meta AI Blog. Retrieved December},
  volume={20},
  pages={2024},
  year={2024}
}

@article{achiam2023gpt,
  title={Gpt-4 technical report},
  author={Achiam, Josh and Adler, Steven and Agarwal, Sandhini and Ahmad, Lama and Akkaya, Ilge and Aleman, Florencia Leoni and Almeida, Diogo and Altenschmidt, Janko and Altman, Sam and Anadkat, Shyamal and others},
  journal={arXiv},
  year={2023}
}

@misc{baror2022sociolinguisticsbasedframeworkstudypoliteness,
      title={Towards a Sociolinguistics-Based Framework for the Study of Politeness in Human-Computer Interaction}, 
      author={Ella Bar-Or and Tom Regev and Paz Shaviv and Noam Tractinsky},
      year={2022},
      eprint={2202.09901},
      archivePrefix={arXiv},
      primaryClass={cs.HC},
      url={https://arxiv.org/abs/2202.09901}, 
}

@article{kearney1984factor,
  title={Factor analysis of the Rathus assertiveness schedule and the personal report of communication apprehension-24: Replication and Extension},
  author={Kearney, Patricia and Beatty, Michael J and Plax, Timothy G and McCroskey, James C},
  journal={Psychological reports},
  volume={54},
  number={3},
  pages={851--854},
  year={1984},
  publisher={SAGE Publications Sage CA: Los Angeles, CA}
}

@article{nevid1979factor,
  title={Factor analysis of the Rathus Assertiveness Schedule with a college population},
  author={Nevid, Jeffrey S and Rathus, Spencer A},
  journal={Journal of Behavior Therapy and Experimental Psychiatry},
  volume={10},
  number={1},
  pages={21--24},
  year={1979},
  publisher={Elsevier}
}

@book{biber1995dimensions,
  title={Dimensions of register variation: A cross-linguistic comparison},
  author={Biber, Douglas},
  year={1995},
  publisher={Cambridge University Press}
}

@article{zou2023representation,
  title={Representation Engineering: A Top-Down Approach to AI Transparency},
  author={Zou, Andy and Phan, Long and Chen, Sarah and Campbell, James and Guo, Phillip and Ren, Richard and Pan, Alexander and Yin, Xuwang and Mazeika, Mantas and Dombrowski, Ann-Kathrin and others},
  journal={CoRR},
  year={2023}
}

@inproceedings{demszky2020goemotions,
  title={GoEmotions: A Dataset of Fine-Grained Emotions},
  author={Demszky, Dorottya and Movshovitz-Attias, Dana and Ko, Jeongwoo and Cowen, Alan and Nemade, Gaurav and Ravi, Sujith},
  booktitle={Proceedings of the 58th Annual Meeting of the Association for Computational Linguistics},
  pages={4040--4054},
  year={2020}
}

@article{zheng2023judging,
  title={Judging llm-as-a-judge with mt-bench and chatbot arena},
  author={Zheng, Lianmin and Chiang, Wei-Lin and Sheng, Ying and Zhuang, Siyuan and Wu, Zhanghao and Zhuang, Yonghao and Lin, Zi and Li, Zhuohan and Li, Dacheng and Xing, Eric and others},
  journal={Advances in neural information processing systems},
  volume={36},
  pages={46595--46623},
  year={2023}
}

@article{pang2019daunting,
  title={The daunting task of real-world textual style transfer auto-evaluation},
  author={Pang, Richard Yuanzhe},
  journal={arXiv preprint arXiv:1910.03747},
  year={2019}
}

@proceedings{emnlp-2014-1,
    title = "Proceedings of the 2014 Conference on Empirical Methods in Natural Language Processing ({EMNLP})",
    editor = "Moschitti, Alessandro  and
      Pang, Bo  and
      Daelemans, Walter",
    month = oct,
    year = "2014",
    address = "Doha, Qatar",
    publisher = "Association for Computational Linguistics",
    url = "https://aclanthology.org/D14-1000/"
}

@article{prabhumoye2018style,
  title={Style transfer through back-translation},
  author={Prabhumoye, Shrimai and Tsvetkov, Yulia and Salakhutdinov, Ruslan and Black, Alan W},
  journal={arXiv preprint arXiv:1804.09000},
  year={2018}
}

@article{shen2017style,
  title={Style transfer from non-parallel text by cross-alignment},
  author={Shen, Tianxiao and Lei, Tao and Barzilay, Regina and Jaakkola, Tommi},
  journal={Advances in neural information processing systems},
  volume={30},
  year={2017}
}

@inproceedings{mukherjee2023low,
  title={Low-resource text style transfer for Bangla: Data \& models},
  author={Mukherjee, Sourabrata and Bansal, Akanksha and Majumdar, Pritha and Ojha, Atul Kr and Du{\v{s}}ek, Ond{\v{r}}ej},
  booktitle={Proceedings of the First Workshop on Bangla Language Processing (BLP-2023)},
  pages={34--47},
  year={2023}
}

@inproceedings{jhamtani2017shakespearizing,
  title={Shakespearizing Modern Language Using Copy-Enriched Sequence to Sequence Models},
  author={Jhamtani, Harsh and Gangal, Varun and Hovy, Eduard and Nyberg, Eric},
  booktitle={Proceedings of the Workshop on Stylistic Variation},
  pages={10--19},
  year={2017}
}

@article{beaglehole2025aggregate,
  title={Aggregate and conquer: detecting and steering LLM concepts by combining nonlinear predictors over multiple layers},
  author={Beaglehole, Daniel and Radhakrishnan, Adityanarayanan and Boix-Adser{\`a}, Enric and Belkin, Mikhail},
  journal={CoRR},
  year={2025}
}

@inproceedings{scalenamulti,
  title={Multi-property Steering of Large Language Models with Dynamic Activation Composition},
  author={Scalena, Daniel and Sarti, Gabriele and Nissim, Malvina},
  booktitle={The 7th BlackboxNLP Workshop-ARR Submissions}
}

@inproceedings{guo2024steering,
  title={Steering Large Language Models for Cross-lingual Information Retrieval},
  author={Guo, Ping and Ren, Yubing and Hu, Yue and Cao, Yanan and Li, Yunpeng and Huang, Heyan},
  booktitle={SIGIR},
  year={2024}
}

@article{wengcontrollm,
  title={ControlLM: Crafting Diverse Personalities for Language Models},
  author={Weng, Yixuan and He, Shizhu and Liu, Kang and Liu, Shengping and Zhao, Jun}
}

@article{zhang2024better,
  title={The better angels of machine personality: How personality relates to llm safety},
  author={Zhang, Jie and Liu, Dongrui and Qian, Chen and Gan, Ziyue and Liu, Yong and Qiao, Yu and Shao, Jing},
  journal={arXiv preprint arXiv:2407.12344},
  year={2024}
}

@article{asghar2016yelp,
  title={Yelp dataset challenge: Review rating prediction},
  author={Asghar, Nabiha},
  journal={arXiv preprint arXiv:1605.05362},
  year={2016}
}

@misc{brooks2024video,
  title={Video generation models as world simulators},
  author={Brooks, Tim and Peebles, Bill and Holmes, Connor and DePue, Will and Guo, Yufei and Jing, Li and Schnurr, David and Taylor, Joe and Luhman, Troy and Luhman, Eric and others},
  howpublished={\href{https://openai. com/research/video-generation-modelsas-world-simulators}{[LINK]}},
  year={2024}
}

@article{brown2020language,
  title={Language models are few-shot learners},
  author={Brown, Tom and Mann, Benjamin and Ryder, Nick and Subbiah, Melanie and Kaplan, Jared D and Dhariwal, Prafulla and Neelakantan, Arvind and Shyam, Pranav and Sastry, Girish and Askell, Amanda and others},
  journal={Advances in neural information processing systems},
  volume={33},
  pages={1877--1901},
  year={2020}
}

@inproceedings{yang2023tailor,
  title={Tailor: A soft-prompt-based approach to attribute-based controlled text generation},
  author={Yang, Kexin and Liu, Dayiheng and Lei, Wenqiang and Yang, Baosong and Xue, Mingfeng and Chen, Boxing and Xie, Jun},
  booktitle={Proceedings of the 61st Annual Meeting of the Association for Computational Linguistics (Volume 1: Long Papers)},
  pages={410--427},
  year={2023}
}

@online{durmus2024persuasion,
author = {Esin Durmus and Liane Lovitt and Alex Tamkin and Stuart Ritchie and Jack Clark and Deep Ganguli},
title = {Measuring the Persuasiveness of Language Models},
date = {2024-04-09},
year = {2024},
url = {https://www.anthropic.com/news/measuring-model-persuasiveness},
}

@article{team2025gemma,
  title={Gemma 3 technical report},
  author={Team, Gemma and Kamath, Aishwarya and Ferret, Johan and Pathak, Shreya and Vieillard, Nino and Merhej, Ramona and Perrin, Sarah and Matejovicova, Tatiana and Ram{\'e}, Alexandre and Rivi{\`e}re, Morgane and others},
  journal={arXiv preprint arXiv:2503.19786},
  year={2025}
}

@article{Spearman1904,
  author    = {Charles Spearman},
  title     = {The Proof and Measurement of Association between Two Things},
  journal   = {The American Journal of Psychology},
  year      = {1904},
  volume    = {15},
  number    = {1},
  pages     = {72--101},
  publisher = {University of Illinois Press},
  doi       = {10.2307/1412159},
  url       = {https://www.jstor.org/stable/1412159}
}

@inproceedings{caa,
    title = "Steering Llama 2 via Contrastive Activation Addition",
    author = "Rimsky, Nina  and
      Gabrieli, Nick  and
      Schulz, Julian  and
      Tong, Meg  and
      Hubinger, Evan  and
      Turner, Alexander",
    editor = "Ku, Lun-Wei  and
      Martins, Andre  and
      Srikumar, Vivek",
    booktitle = "Proceedings of the 62nd Annual Meeting of the Association for Computational Linguistics (Volume 1: Long Papers)",
    month = aug,
    year = "2024",
    address = "Bangkok, Thailand",
    publisher = "Association for Computational Linguistics",
    url = "https://aclanthology.org/2024.acl-long.828/",
    doi = "10.18653/v1/2024.acl-long.828",
    pages = "15504--15522"
}

@inproceedings{
pplm,
title={Plug and Play Language Models: A Simple Approach to Controlled Text Generation},
author={Sumanth Dathathri and Andrea Madotto and Janice Lan and Jane Hung and Eric Frank and Piero Molino and Jason Yosinski and Rosanne Liu},
booktitle={International Conference on Learning Representations},
year={2020},
url={https://openreview.net/forum?id=H1edEyBKDS}
}

@inproceedings{gedi,
    title = "{G}e{D}i: Generative Discriminator Guided Sequence Generation",
    author = "Krause, Ben  and
      Gotmare, Akhilesh Deepak  and
      McCann, Bryan  and
      Keskar, Nitish Shirish  and
      Joty, Shafiq  and
      Socher, Richard  and
      Rajani, Nazneen Fatema",
    editor = "Moens, Marie-Francine  and
      Huang, Xuanjing  and
      Specia, Lucia  and
      Yih, Scott Wen-tau",
    booktitle = "Findings of the Association for Computational Linguistics: EMNLP 2021",
    month = nov,
    year = "2021",
    address = "Punta Cana, Dominican Republic",
    publisher = "Association for Computational Linguistics",
    url = "https://aclanthology.org/2021.findings-emnlp.424/",
    doi = "10.18653/v1/2021.findings-emnlp.424",
    pages = "4929--4952"
}

@inproceedings{fudge,
    title = "{FUDGE}: Controlled Text Generation With Future Discriminators",
    author = "Yang, Kevin  and
      Klein, Dan",
    editor = "Toutanova, Kristina  and
      Rumshisky, Anna  and
      Zettlemoyer, Luke  and
      Hakkani-Tur, Dilek  and
      Beltagy, Iz  and
      Bethard, Steven  and
      Cotterell, Ryan  and
      Chakraborty, Tanmoy  and
      Zhou, Yichao",
    booktitle = "Proceedings of the 2021 Conference of the North American Chapter of the Association for Computational Linguistics: Human Language Technologies",
    month = jun,
    year = "2021",
    address = "Online",
    publisher = "Association for Computational Linguistics",
    url = "https://aclanthology.org/2021.naacl-main.276/",
    doi = "10.18653/v1/2021.naacl-main.276",
    pages = "3511--3535"
}

@inproceedings{
cold,
title={{COLD} Decoding: Energy-based Constrained Text Generation with Langevin Dynamics},
author={Lianhui Qin and Sean Welleck and Daniel Khashabi and Yejin Choi},
booktitle={Advances in Neural Information Processing Systems},
editor={Alice H. Oh and Alekh Agarwal and Danielle Belgrave and Kyunghyun Cho},
year={2022},
url={https://openreview.net/forum?id=TiZYrQ-mPup}
}

@inproceedings{bolt,
    title = "{BOLT}: Fast Energy-based Controlled Text Generation with Tunable Biases",
    author = "Liu, Xin  and
      Khalifa, Muhammad  and
      Wang, Lu",
    editor = "Rogers, Anna  and
      Boyd-Graber, Jordan  and
      Okazaki, Naoaki",
    booktitle = "Proceedings of the 61st Annual Meeting of the Association for Computational Linguistics (Volume 2: Short Papers)",
    month = jul,
    year = "2023",
    address = "Toronto, Canada",
    publisher = "Association for Computational Linguistics",
    url = "https://aclanthology.org/2023.acl-short.18/",
    doi = "10.18653/v1/2023.acl-short.18",
    pages = "186--200"
}

@inproceedings{dexperts,
    title = "{DE}xperts: Decoding-Time Controlled Text Generation with Experts and Anti-Experts",
    author = "Liu, Alisa  and
      Sap, Maarten  and
      Lu, Ximing  and
      Swayamdipta, Swabha  and
      Bhagavatula, Chandra  and
      Smith, Noah A.  and
      Choi, Yejin",
    editor = "Zong, Chengqing  and
      Xia, Fei  and
      Li, Wenjie  and
      Navigli, Roberto",
    booktitle = "Proceedings of the 59th Annual Meeting of the Association for Computational Linguistics and the 11th International Joint Conference on Natural Language Processing (Volume 1: Long Papers)",
    month = aug,
    year = "2021",
    address = "Online",
    publisher = "Association for Computational Linguistics",
    url = "https://aclanthology.org/2021.acl-long.522/",
    doi = "10.18653/v1/2021.acl-long.522",
    pages = "6691--6706"
}

@inproceedings{steerlm,
    title = "{S}teer{LM}: Attribute Conditioned {SFT} as an (User-Steerable) Alternative to {RLHF}",
    author = "Dong, Yi  and
      Wang, Zhilin  and
      Sreedhar, Makesh  and
      Wu, Xianchao  and
      Kuchaiev, Oleksii",
    editor = "Bouamor, Houda  and
      Pino, Juan  and
      Bali, Kalika",
    booktitle = "Findings of the Association for Computational Linguistics: EMNLP 2023",
    month = dec,
    year = "2023",
    address = "Singapore",
    publisher = "Association for Computational Linguistics",
    url = "https://aclanthology.org/2023.findings-emnlp.754/",
    doi = "10.18653/v1/2023.findings-emnlp.754",
    pages = "11275--11288"
}

@inproceedings{sctg,
    title = "Evaluating the Smooth Control of Attribute Intensity in Text Generation with {LLM}s",
    author = "Zhou, Shang  and
      Yao, Feng  and
      Dong, Chengyu  and
      Wang, Zihan  and
      Shang, Jingbo",
    editor = "Ku, Lun-Wei  and
      Martins, Andre  and
      Srikumar, Vivek",
    booktitle = "Findings of the Association for Computational Linguistics: ACL 2024",
    month = aug,
    year = "2024",
    address = "Bangkok, Thailand",
    publisher = "Association for Computational Linguistics",
    url = "https://aclanthology.org/2024.findings-acl.258/",
    doi = "10.18653/v1/2024.findings-acl.258",
    pages = "4348--4362"
}

@inproceedings{
wu2025axbench,
title={AxBench: Steering {LLM}s? Even Simple Baselines Outperform Sparse Autoencoders},
author={Zhengxuan Wu and Aryaman Arora and Atticus Geiger and Zheng Wang and Jing Huang and Dan Jurafsky and Christopher D Manning and Christopher Potts},
booktitle={Forty-second International Conference on Machine Learning},
year={2025},
url={https://openreview.net/forum?id=K2CckZjNy0}
}

@article{qwen3,
    title={Qwen3 Technical Report}, 
    author={An Yang and Anfeng Li and Baosong Yang and Beichen Zhang and Binyuan Hui and Bo Zheng and Bowen Yu and Chang Gao and Chengen Huang and Chenxu Lv and Chujie Zheng and Dayiheng Liu and Fan Zhou and Fei Huang and Feng Hu and Hao Ge and Haoran Wei and Huan Lin and Jialong Tang and Jian Yang and Jianhong Tu and Jianwei Zhang and Jianxin Yang and Jiaxi Yang and Jing Zhou and Jingren Zhou and Junyang Lin and Kai Dang and Keqin Bao and Kexin Yang and Le Yu and Lianghao Deng and Mei Li and Mingfeng Xue and Mingze Li and Pei Zhang and Peng Wang and Qin Zhu and Rui Men and Ruize Gao and Shixuan Liu and Shuang Luo and Tianhao Li and Tianyi Tang and Wenbiao Yin and Xingzhang Ren and Xinyu Wang and Xinyu Zhang and Xuancheng Ren and Yang Fan and Yang Su and Yichang Zhang and Yinger Zhang and Yu Wan and Yuqiong Liu and Zekun Wang and Zeyu Cui and Zhenru Zhang and Zhipeng Zhou and Zihan Qiu},
    journal = {arXiv preprint arXiv:2505.09388},
    year={2025}
}

@article{openai2023gpt,
  title={Gpt-4 technical report. arxiv 2303.08774},
  author={OpenAI, R},
  journal={View in Article},
  volume={2},
  number={5},
  pages={1},
  year={2023}
}

@article{fisher1915frequency,
  title        = {Frequency Distribution of the Values of the Correlation Coefficient in Samples from an Indefinitely Large Population},
  author       = {Fisher, Ronald A.},
  journal      = {Biometrika},
  volume       = {10},
  number       = {4},
  pages        = {507--521},
  year         = {1915},
  doi          = {10.1093/biomet/10.4.507}
}

\newpage
\appendix

\clearpage
\twocolumn

\section{Risks and Ethics}

There are no risks or ethical concerns with this work.

\section{Licensing}
This work is conducted on datasets that are either publicly available or authorized for research use.  We ensured that the use of all existing datasets was consistent with their original intended use as specified by their licenses. Similarly, all models used in experiments in this work were used as dictated by their respective licenses. We used Co-pilot and AI Assistants to support human-generated artifacts.
\section{Extended Related Work}

\paragraph{Prompting for Concept Control.} 
Prompt-based methods, including prefix-tuning, soft prompts, and learned prompt vectors, have emerged as lightweight alternatives to full model fine-tuning for controllable text generation. Prefix-tuning has been used to inject attributes without retraining the model~\cite{ref1,ref2}, extended to multi-aspect settings through plugin modules and disentanglement objectives~\cite{ref4,ref10}. Other approaches learn attribute-specific soft prompts, either with contrastive training~\cite{ref12}, latent prior manipulation~\cite{ref13}, or interference-reducing designs such as Tailor~\cite{ref14}. DisCup~\cite{ref20} further integrates discriminator feedback into prompt learning, while Attribute Alignment~\cite{ref31} builds on conditioning mechanisms. These methods show strong controllability but require training effort and often struggle to generalize across multiple attributes.

\paragraph{Representation Engineering and Steering.}
Representation engineering (RepE) methods manipulate hidden activations to steer model behavior. They have been shown effective in controlling sentiment (e.g., shifting polarity or tone)~\cite{turnersteering,konen2024style,cai2024self,zou2023representation}, typically using datasets such as GoEmotions~\cite{demszky2020goemotions} or Yelp~\cite{asghar2016yelp}. Beyond sentiment, RepE has been extended to personality traits, steering along MBTI~\cite{zhang2024better} or OCEAN~\cite{wengcontrollm} dimensions, influencing reasoning style, honesty, and conversational stance. Other work explores steering for language, style, and genre, including cross-lingual transfer~\cite{guo2024steering,scalenamulti}, or stylized generation~\cite{konen2024style,beaglehole2025aggregate}. Recent steering techniques such as Contrastive Activation Addition (CAA)~\cite{caa} provide training-free, intensity-scalable control vectors derived from positive/negative exemplars, aligning closely with the idea of numeric sliders. However, most RepE studies focus on one attribute at a time, without probing how multiple steering directions interact. Further to this, \citet{wu2025axbench} demonstrate that simple prompting often performs much better than many of the more complex RepE methods discussed above.

\paragraph{Style Transfer and Multi-Attribute Control.}
Supervised text style transfer methods rely on parallel corpora and sequence-to-sequence models~\cite{jhamtani2017shakespearizing,mukherjee2023low}, but are constrained by scarce paired data. Unsupervised methods for non-parallel data include prototype editing (swapping style markers with target-style phrases)~\cite{mukherjee2023low}, or disentanglement strategies that factorize semantics and style, recombining them via back-translation or adversarial training~\cite{shen2017style,prabhumoye2018style}. While effective for coarse style shifts, these approaches are not naturally suited for fine-grained numeric control or multi-attribute specification.

\paragraph{Fine-Grained Control: Single-Attribute Methods.}
Most work on fine-grained control introduces a continuous “knob” for a \emph{single} attribute, with evaluation focused on calibration along that one dimension. Families include:
\begin{itemize}
    \item \textit{Decoding-time guidance.} PPLM~\cite{pplm} backpropagates from an attribute classifier through LM hidden states at generation time; GeDi~\cite{gedi} trains small conditional LMs to reweight token probabilities; FUDGE~\cite{fudge} trains discriminators predicting sequence-level attributes from partial prefixes; and energy/logit methods such as COLD~\cite{cold} and BOLT~\cite{bolt} add attribute-specific energies or biases. Each provides a tunable weight parameter, enabling smooth control of attribute intensity.
    \item \textit{Product-of-experts.} DExperts~\cite{dexperts} combine base LMs with expert/anti-expert models, where the mixture coefficient $\alpha$ controls strength.
    \item \textit{Activation steering.} Contrastive Activation Addition (CAA)~\cite{caa} computes steering directions from exemplar differences and scales them at inference.
    \item \textit{Training-time numeric control.} SteerLM~\cite{steerlm} finetunes on data labeled with regressor-predicted attribute values, allowing users to set numeric controls such as “positivity=7/10.” This achieves excellent calibration but requires labeled data and SFT cycles.
\end{itemize}
These methods demonstrate smooth, single-attribute control, but rarely extend to \emph{dual-concept} settings. While some (e.g., FUDGE, GeDi, or energy-based methods) can in principle compose multiple guidance signals by assigning separate weights $\lambda_a, \lambda_b$, systematic evaluation of interference between attributes remains limited.

\paragraph{Towards Multi-Concept Fine-Grained Control.}
Recent frameworks begin to explore smooth, fine-grained evaluation. The Smoothly Controllable Text Generation (SCTG) benchmark~\cite{sctg} defines fine-grained control as the ability to vary an attribute over a 10-point scale, using LLM-as-judge with Elo-style pairwise comparisons to assess calibration and relevance. However, SCTG focuses exclusively on single-attribute scenarios. In contrast, our work explicitly evaluates \emph{dual-concept} fine-grained control, introducing systematic protocols to measure interference when varying one concept while holding another fixed. This perspective highlights the challenges of compositional control and the need for methods robust to attribute entanglement.

\paragraph{Evaluation of Controllability.}
Evaluation typically relies on automatic classifiers trained to predict style or attribute labels on generated outputs~\cite{emnlp-2014-1}. While efficient, such classifiers often suffer from subjectivity and domain mismatch~\cite{pang2019daunting}. Human evaluation remains the gold standard but is costly and inconsistent. More recent work explores LLMs themselves as judges~\cite{zheng2023judging,sun2023evaluating}, providing scalable and flexible evaluation pipelines. Our evaluation setup builds on this line, using pairwise comparisons with strong judge models to assess fine-grained controllability in both single- and dual-concept scenarios.

\onecolumn
\newpage
\section{Detailed Results}
\label{app:allresults}

\begin{table}[htb!]
\centering
\small
\setlength{\tabcolsep}{5pt}
\begin{tabular}{l*{6}{c}}
\toprule
& \textbf{Llama-3B} & \textbf{Llama-11B} 
& \textbf{Gemma-4B} & \textbf{Gemma-12B} 
& \textbf{Phi-3.8B} & \textbf{Qwen-14B}\\
\midrule
$C_a$ (single)          & $-0.02{\sd{0.52}}$ & $-0.02{\sd{0.52}}$ & $-0.33{\sd{0.47}}$ & $0.52{\sd{0.46}}$ & $-0.05{\sd{0.54}}$ & $0.65{\sd{0.30}}$ \\
$C_a \mid (C_b = 0)$    & $-0.11{\sd{0.53}}$ & $0.12{\sd{0.51}}$ & $0.15{\sd{0.54}}$ & $0.24{\sd{0.48}}$ & $0.08{\sd{0.46}}$ & $0.70{\sd{0.28}}$\\
$C_a \mid (C_b = 1)$    & $-0.12{\sd{0.49}}$ & $-0.07{\sd{0.53}}$ & $0.12{\sd{0.47}}$ & $0.01{\sd{0.55}}$ & $-0.06{\sd{0.48}}$ & $0.72{\sd{0.26}}$\\
$C_a \mid (C_b = 2)$    & $0.01{\sd{0.48}}$ & $0.03{\sd{0.54}}$ & $0.16{\sd{0.45}}$ & $0.03{\sd{0.58}}$ & $-0.19{\sd{0.46}}$ & $0.62{\sd{0.35}}$\\
$C_a \mid (C_b = 3)$    & $-0.09{\sd{0.49}}$ & $0.07{\sd{0.53}}$ & $-0.08{\sd{0.47}}$ & $-0.02{\sd{0.58}}$ & $-0.07{\sd{0.47}}$ & $0.61{\sd{0.40}}$\\
$C_a \mid (C_b = 4)$    & $0.15{\sd{0.48}}$ & $-0.03{\sd{0.53}}$ & $-0.215{\sd{0.52}}$ & $-0.16{\sd{0.53}}$ & $-0.24{\sd{0.48}}$ & $0.55{\sd{0.36}}$\\
$C_a \mid C_b$ fixed    & $-0.03{\sd{0.50}}$ & $0.02{\sd{0.53}}$ & $0.03{\sd{0.51}}$ & $0.02{\sd{0.56}}$ & $-0.09{\sd{0.48}}$ & $0.64{\sd{0.34}}$\\
\bottomrule
\end{tabular}
\caption{ARGUMENT generation \textbf{clarity--politeness}.}
\label{tab:arg-clar-polit}
\end{table}

\begin{table}[htb!]
\centering
\small
\setlength{\tabcolsep}{5pt}
\begin{tabular}{l*{6}{c}}
\toprule
& \textbf{Llama-3B} & \textbf{Llama-11B} 
& \textbf{Gemma-4B} & \textbf{Gemma-12B} 
& \textbf{Phi-3.8B} & \textbf{Qwen-14B}\\
\midrule
$C_a$ (single)          & $0.77{\sd{0.27}}$ & $0.92{\sd{0.09}}$ & $0.94{\sd{0.09}}$ & $0.98{\sd{0.03}}$ & $0.69{\sd{0.28}}$ & $0.99{\sd{0.02}}$\\
$C_a \mid (C_b = 0)$    & $0.34{\sd{0.47}}$ & $0.33{\sd{0.40}}$ & $0.91{\sd{0.12}}$ & $0.97{\sd{0.05}}$ & $0.44{\sd{0.41}}$ & $0.95{\sd{0.08}}$\\
$C_a \mid (C_b = 1)$    & $0.41{\sd{0.39}}$ & $0.66{\sd{0.36}}$ & $0.90{\sd{0.13}}$ & $0.97{\sd{0.06}}$ & $0.38{\sd{0.41}}$ & $0.98{\sd{0.04}}$\\
$C_a \mid (C_b = 2)$    & $0.44{\sd{0.47}}$ & $0.66{\sd{0.32}}$ & $0.87{\sd{0.16}}$ & $0.97{\sd{0.04}}$ & $0.49{\sd{0.38}}$ & $0.98{\sd{0.04}}$\\
$C_a \mid (C_b = 3)$    & $0.40{\sd{0.45}}$ & $0.71{\sd{0.27}}$ & $0.84{\sd{0.21}}$ & $0.96{\sd{0.05}}$ & $0.51{\sd{0.37}}$ & $0.98{\sd{0.03}}$\\
$C_a \mid (C_b = 4)$    & $0.36{\sd{0.43}}$ & $0.43{\sd{0.46}}$ & $0.88{\sd{0.17}}$ & $0.96{\sd{0.06}}$ & $0.58{\sd{0.37}}$ & $0.98{\sd{0.03}}$\\
$C_a \mid C_b$ fixed    & $0.39{\sd{0.44}}$ & $0.56{\sd{0.40}}$ & $0.88{\sd{0.16}}$ & $0.97{\sd{0.05}}$ & $0.48{\sd{0.41}}$ & $0.97{\sd{0.05}}$\\
\bottomrule
\end{tabular}
\caption{ARGUMENT generation \textbf{formality--assertiveness}.}
\label{tab:arg-form-assert}
\end{table}

\begin{table}[htb!]
\centering
\small
\setlength{\tabcolsep}{5pt}
\begin{tabular}{l*{6}{c}}
\toprule
& \textbf{Llama-3B} & \textbf{Llama-11B} 
& \textbf{Gemma-4B} & \textbf{Gemma-12B} 
& \textbf{Phi-3.8B} & \textbf{Qwen-14B}\\
\midrule
$C_a$ (single)          & $0.46{\sd{0.40}}$ & $0.76{\sd{0.23}}$ & $0.65{\sd{0.32}}$ & $0.95{\sd{0.07}}$ & $0.38{\sd{0.43}}$ & $0.92{\sd{0.11}}$\\
$C_a \mid (C_b = 0)$    & $0.27{\sd{0.51}}$ & $0.32{\sd{0.46}}$ & $0.38{\sd{0.53}}$ & $0.89{\sd{0.14}}$ & $0.15{\sd{0.54}}$ & $0.91{\sd{0.11}}$\\
$C_a \mid (C_b = 1)$    & $0.18{\sd{0.47}}$ & $0.52{\sd{0.39}}$ & $0.28{\sd{0.49}}$ & $0.84{\sd{0.18}}$ & $0.16{\sd{0.44}}$ & $0.90{\sd{0.15}}$\\
$C_a \mid (C_b = 2)$    & $0.13{\sd{0.49}}$ & $0.47{\sd{0.41}}$ & $0.36{\sd{0.40}}$ & $0.87{\sd{0.17}}$ & $0.18{\sd{0.47}}$ & $0.86{\sd{0.14}}$\\
$C_a \mid (C_b = 3)$    & $0.18{\sd{0.51}}$ & $0.64{\sd{0.35}}$ & $0.35{\sd{0.46}}$ & $0.89{\sd{0.11}}$ & $0.21{\sd{0.45}}$ & $0.86{\sd{0.19}}$\\
$C_a \mid (C_b = 4)$    & $0.29{\sd{0.49}}$ & $0.60{\sd{0.37}}$ & $0.53{\sd{0.39}}$ & $0.92{\sd{0.09}}$ & $0.10{\sd{0.53}}$ & $0.88{\sd{0.14}}$\\
$C_a \mid C_b$ fixed    & $0.21{\sd{0.50}}$ & $0.51{\sd{0.41}}$ & $0.38{\sd{0.46}}$ & $0.88{\sd{0.14}}$ & $0.16{\sd{0.49}}$ & $0.88{\sd{0.15}}$\\
\bottomrule
\end{tabular}
\caption{ARGUMENT generation \textbf{humor--persuasiveness}.}
\label{tab:arg-humor-pers}
\end{table}


\begin{table}[htb!]
\centering
\small
\setlength{\tabcolsep}{5pt}
\begin{tabular}{l*{6}{c}}
\toprule
& \textbf{Llama-3B} & \textbf{Llama-11B} 
& \textbf{Gemma-4B} & \textbf{Gemma-12B} 
& \textbf{Phi-3.8B} & \textbf{Qwen-14B} \\
\midrule
$C_a$ (single)          & $0.20{\sd{0.54}}$ & $0.45{\sd{0.46}}$ & $0.22{\sd{0.53}}$ & $0.92{\sd{0.11}}$ & $0.01{\sd{0.52}}$ & $0.89{\sd{0.12}}$\\
$C_a \mid (C_b = 0)$    & $-0.08{\sd{0.52}}$ & $-0.10{\sd{0.51}}$ & $-0.05{\sd{0.55}}$ & $0.37{\sd{0.30}}$ & $-0.08{\sd{0.52}}$ & $0.65{\sd{0.34}}$\\
$C_a \mid (C_b = 1)$    & $0.05{\sd{0.52}}$ & $-0.02{\sd{0.46}}$ & $-0.10{\sd{0.49}}$ & $0.35{\sd{0.48}}$ & $-0.14{\sd{0.48}}$ & $0.68{\sd{0.28}}$ \\
$C_a \mid (C_b = 2)$    & $-0.01{\sd{0.55}}$ & $-0.02{\sd{59}}$ & $-0.09{\sd{0.54}}$ & $0.38{\sd{0.43}}$ & $-0.04{\sd{0.50}}$ & $0.80{\sd{0.20}}$\\
$C_a \mid (C_b = 3)$    & $0.05{\sd{0.53}}$ & $0.03{\sd{0.49}}$ & $0.02{\sd{0.48}}$ & $0.39{\sd{0.46}}$ & $0.06{\sd{0.45}}$ & $0.78{\sd{0.17}}$\\
$C_a \mid (C_b = 4)$    & $0.14{\sd{0.47}}$ & $0.05{\sd{0.56}}$ & $-0.06{\sd{0.52}}$ & $0.29{\sd{0.45}}$ & $0.02{\sd{0.53}}$ & $0.77{\sd{0.22}}$\\
$C_a \mid C_b$ fixed    & $0.03{\sd{0.52}}$ & $-0.01{\sd{0.53}}$ & $-0.06{\sd{0.52}}$ & $0.35{\sd{0.43}}$ & $-0.04{\sd{0.50}}$ & $0.74{\sd{0.26}}$\\
\bottomrule
\end{tabular}
\caption{STORY generation \textbf{clarity--politeness}.}
\label{tab:story-clar-polit}
\end{table}

\begin{table}[htb!]
\centering
\small
\setlength{\tabcolsep}{5pt}
\begin{tabular}{l*{6}{c}}
\toprule
& \textbf{Llama-3B} & \textbf{Llama-11B} 
& \textbf{Gemma-4B} & \textbf{Gemma-12B} 
& \textbf{Phi-3.8B} & \textbf{Qwen-14B} \\
\midrule
$C_a$ (single)          & $0.85{\sd{0.18}}$ & $0.93{\sd{0.09}}$ & $0.83{\sd{0.21}}$ & $1.00{\sd{0.02}}$ & $0.53{\sd{0.39}}$ & $0.98{\sd{0.05}}$\\
$C_a \mid (C_b = 0)$    & $0.72{\sd{0.27}}$ & $0.66{\sd{0.31}}$ & $0.89{\sd{0.14}}$ & $0.98{\sd{0.05}}$ & $0.36{\sd{0.48}}$ & $0.96{\sd{0.06}}$\\
$C_a \mid (C_b = 1)$    & $0.79{\sd{0.28}}$ & $0.83{\sd{0.21}}$ & $0.93{\sd{0.10}}$ & $0.98{\sd{0.04}}$ & $0.40{\sd{0.39}}$ & $0.97{\sd{0.05}}$\\
$C_a \mid (C_b = 2)$    & $0.75{\sd{0.25}}$ & $0.81{\sd{0.21}}$ & $0.87{\sd{0.15}}$ & $0.95{\sd{0.07}}$ & $0.21{\sd{0.48}}$ & $0.98{\sd{0.04}}$ \\
$C_a \mid (C_b = 3)$    & $0.83{\sd{0.19}}$ & $0.79{\sd{0.20}}$ & $0.80{\sd{0.19}}$ & $0.96{\sd{0.07}}$ & $0.37{\sd{0.39}}$ & $0.98{\sd{0.04}}$\\
$C_a \mid (C_b = 4)$    & $0.83{\sd{0.21}}$ & $0.79{\sd{0.26}}$ & $0.76{\sd{0.25}}$ & $0.95{\sd{0.11}}$ & $0.39{\sd{0.47}}$ & $0.98{\sd{0.04}}$\\
$C_a \mid C_b$ fixed    & $0.78{\sd{0.25}}$ & $0.77{\sd{0.25}}$ & $0.85{\sd{0.18}}$ & $0.96{\sd{0.07}}$ & $0.35{\sd{0.45}}$ & $0.97{\sd{0.05}}$\\
\bottomrule
\end{tabular}
\caption{STORY generation \textbf{formality--assertiveness}.}
\label{tab:story-form-assert}
\end{table}

\begin{table}[htb!]
\centering
\small
\setlength{\tabcolsep}{5pt}
\begin{tabular}{l*{6}{c}}
\toprule
& \textbf{Llama-3B} & \textbf{Llama-11B} 
& \textbf{Gemma-4B} & \textbf{Gemma-12B} 
& \textbf{Phi-3.8B} & \textbf{Qwen-14B}\\
\midrule
$C_a$ (single)          & $0.30{\sd{0.50}}$ & $0.81{\sd{0.26}}$ & $0.57{\sd{0.41}}$ & $0.95{\sd{0.06}}$ & $0.17{\sd{0.51}}$ & $0.92{\sd{0.10}}$\\
$C_a \mid (C_b = 0)$    & $0.16{\sd{0.49}}$ & $0.24{\sd{0.42}}$ & $0.38{\sd{0.45}}$ & $0.87{\sd{0.17}}$ & $0.03{\sd{0.48}}$ & $0.93{\sd{0.10}}$\\
$C_a \mid (C_b = 1)$    & $0.15{\sd{0.56}}$ & $0.43{\sd{0.46}}$ & $0.27{\sd{0.43}}$ & $0.84{\sd{0.18}}$ & $0.06{\sd{0.49}}$ & $0.91{\sd{0.12}}$\\
$C_a \mid (C_b = 2)$    & $0.12{\sd{0.53}}$ & $0.40{\sd{0.41}}$ & $0.43{\sd{0.41}}$ & $0.81{\sd{0.24}}$ & $0.02{\sd{0.53}}$ & $0.90{\sd{0.11}}$\\
$C_a \mid (C_b = 3)$    & $0.12{\sd{0.50}}$ & $0.28{\sd{0.46}}$ & $0.16{\sd{0.50}}$ & $0.77{\sd{0.25}}$ & $-0.01{\sd{0.43}}$ & $0.89{\sd{0.11}}$\\
$C_a \mid (C_b = 4)$    & $0.19{\sd{0.51}}$ & $0.47{\sd{0.44}}$ & $0.22{\sd{0.52}}$ & $0.77{\sd{0.22}}$ & $-0.01{\sd{0.58}}$ & $0.88{\sd{0.14}}$\\
$C_a \mid C_b$ fixed    & $0.15{\sd{0.52}}$ & $0.36{\sd{0.45}}$ & $0.29{\sd{0.47}}$ & $0.81{\sd{0.22}}$ & $0.02{\sd{0.50}}$ & $0.90{\sd{0.12}}$\\
\bottomrule
\end{tabular}
\caption{STORY generation \textbf{humor--persuasiveness}.}
\label{tab:story-humor-pers}
\end{table}

\begin{table}[htb!]
\centering
\small
\setlength{\tabcolsep}{5pt}
\begin{tabular}{l*{6}{c}}
\toprule
& \textbf{Llama-3B} & \textbf{Llama-11B} 
& \textbf{Gemma-4B} & \textbf{Gemma-12B} 
& \textbf{Phi-3.8B} & \textbf{Qwen-14B}\\
\midrule
$C_a$ (single)          & $0.06{\sd{0.64}}$ & $0.21{\sd{0.56}}$ & $0.13{\sd{0.64}}$ & $0.15{\sd{0.61}}$ & $0.03{\sd{0.53}}$ & $0.64{\sd{0.21}}$\\
$C_a \mid (C_b = 0)$    & $-0.14{\sd{0.48}}$ & $-0.01{\sd{0.42}}$ & $0.07{\sd{0.40}}$ & $0.07{\sd{0.44}}$ & $-0.13{\sd{0.50}}$ & $0.31{\sd{0.40}}$\\
$C_a \mid (C_b = 1)$    & $-0.09{\sd{0.43}}$ & $-0.02{\sd{0.46}}$ & $0.02{\sd{0.37}}$ & $-0.12{\sd{0.50}}$ & $-0.09{\sd{0.50}}$ & $0.42{\sd{0.38}}$\\
$C_a \mid (C_b = 2)$    & $-0.04{\sd{0.38}}$ & $-0.06{\sd{0.45}}$ & $0.00{\sd{0.41}}$ & $-0.40{\sd{0.53}}$ & $0.01{\sd{0.45}}$ & $0.46{\sd{0.40}}$\\
$C_a \mid (C_b = 3)$    & $0.04{\sd{0.37}}$ & $0.08{\sd{0.44}}$ & $0.00{\sd{0.47}}$ & $-0.35{\sd{0.53}}$ & $-0.13{\sd{0.49}}$ & $0.33{\sd{0.50}}$\\
$C_a \mid (C_b = 4)$    & $0.02{\sd{0.41}}$ & $0.12{\sd{0.45}}$ & $0.03{\sd{0.57}}$ & $-0.44{\sd{0.46}}$ & $-0.04{\sd{0.50}}$ & $0.43{\sd{0.45}}$\\
$C_a \mid C_b$ fixed    & $-0.04{\sd{0.42}}$ & $0.02{\sd{0.45}}$ & $0.02{\sd{0.45}}$ & $-0.25{\sd{0.53}}$ & $-0.07{\sd{0.49}}$ & $0.39{\sd{0.43}}$\\
\bottomrule
\end{tabular}
\caption{STRUCTURED generation \textbf{clarity--politeness}.}
\label{tab:clar-polit-full}
\end{table}

\begin{table}[htb!]
\centering
\small
\setlength{\tabcolsep}{5pt}
\begin{tabular}{l*{6}{c}}
\toprule
& \textbf{Llama-3B} & \textbf{Llama-11B} 
& \textbf{Gemma-4B} & \textbf{Gemma-12B} 
& \textbf{Phi-3.8B} & \textbf{Qwen-14B}\\
\midrule
$C_a$ (single)          & $0.59{\sd{0.48}}$ & $0.80{\sd{0.24}}$ & $0.90{\sd{0.22}}$ & $0.93{\sd{0.14}}$ & $0.73{\sd{0.26}}$ & $0.96{\sd{0.07}}$\\
$C_a \mid (C_b = 0)$    & $0.62{\sd{0.39}}$ & $0.18{\sd{0.48}}$ & $0.11{\sd{0.50}}$ & $0.61{\sd{0.44}}$ & $0.31{\sd{0.44}}$ & $0.78{\sd{0.22}}$\\
$C_a \mid (C_b = 1)$    & $0.66{\sd{0.32}}$ & $0.42{\sd{0.39}}$ & $0.10{\sd{0.44}}$ & $0.71{\sd{0.35}}$ & $0.46{\sd{0.41}}$ & $0.88{\sd{0.13}}$\\
$C_a \mid (C_b = 2)$    & $0.69{\sd{0.40}}$ & $0.47{\sd{0.45}}$ & $0.28{\sd{0.44}}$ & $0.79{\sd{0.32}}$ & $0.43{\sd{0.49}}$ & $0.92{\sd{0.14}}$\\
$C_a \mid (C_b = 3)$    & $0.78{\sd{0.33}}$ & $0.52{\sd{0.43}}$ & $0.44{\sd{0.41}}$ & $0.88{\sd{0.18}}$ & $0.35{\sd{0.41}}$ & $0.91{\sd{0.11}}$\\
$C_a \mid (C_b = 4)$    & $0.85{\sd{0.21}}$ & $0.51{\sd{0.40}}$ & $0.55{\sd{0.38}}$ & $0.86{\sd{0.19}}$ & $0.47{\sd{0.40}}$ & $0.90{\sd{0.07}}$\\
$C_a \mid C_b$ fixed    & $0.72{\sd{0.35}}$ & $0.42{\sd{0.45}}$ & $0.30{\sd{0.47}}$ & $0.77{\sd{0.33}}$ & $0.40{\sd{0.44}}$ & $0.88{\sd{0.15}}$\\
\bottomrule
\end{tabular}
\caption{STRUCTURED generation \textbf{formality--assertiveness}.}
\label{tab:form-assert-full}
\end{table}

\begin{table}[htb!]
\centering
\small
\setlength{\tabcolsep}{5pt}
\begin{tabular}{l*{6}{c}}
\toprule
& \textbf{Llama-3B} & \textbf{Llama-11B} 
& \textbf{Gemma-4B} & \textbf{Gemma-12B} 
& \textbf{Phi-3.8B} & \textbf{Qwen-14B}\\
\midrule
$C_a$ (single)          & $0.47{\sd{0.44}}$ & $0.73{\sd{0.22}}$ & $0.86{\sd{0.12}}$ & $0.94{\sd{0.12}}$ & $0.54{\sd{0.38}}$ & $0.90{\sd{0.12}}$ \\
$C_a \mid (C_b = 0)$    & $0.33{\sd{0.53}}$ & $0.19{\sd{0.52}}$ & $0.42{\sd{0.41}}$ & $0.89{\sd{0.15}}$ & $0.11{\sd{0.52}}$ & $0.85{\sd{0.21}}$ \\
$C_a \mid (C_b = 1)$    & $0.23{\sd{0.56}}$ & $0.28{\sd{0.48}}$ & $0.22{\sd{0.48}}$ & $0.88{\sd{0.17}}$ & $0.09{\sd{0.53}}$ & $0.86{\sd{0.17}}$\\
$C_a \mid (C_b = 2)$    & $0.21{\sd{0.50}}$ & $0.31{\sd{0.52}}$ & $0.36{\sd{0.45}}$ & $0.86{\sd{0.16}}$ & $0.18{\sd{0.45}}$ & $0.80{\sd{0.20}}$\\
$C_a \mid (C_b = 3)$    & $0.27{\sd{0.52}}$ & $0.36{\sd{0.48}}$ & $0.45{\sd{0.43}}$ & $0.88{\sd{0.12}}$ & $0.34{\sd{0.40}}$ & $0.88{\sd{0.12}}$\\
$C_a \mid (C_b = 4)$    & $0.14{\sd{0.50}}$ & $0.44{\sd{0.49}}$ & $0.52{\sd{0.38}}$ & $0.91{\sd{0.12}}$ & $0.21{\sd{0.46}}$ & $0.83{\sd{0.22}}$\\
$C_a \mid C_b$ fixed    & $0.24{\sd{0.53}}$ & $0.31{\sd{0.50}}$ & $0.39{\sd{0.44}}$ & $0.88{\sd{0.15}}$ & $0.19{\sd{0.48}}$ & $0.84{\sd{0.19}}$\\
\bottomrule
\end{tabular}
\caption{STRUCTURED generation \textbf{humour--persuasiveness}.}
\label{tab:humour-pers-full}
\end{table}
\onecolumn
\newpage
\section{Detailed Fisher Score Results}
\label{app:fisherresults}
To mitigate the nonlinearity and boundedness of correlation coefficients when aggregating across contexts, we also report results after applying Fisher’s $z$-transformation to the Spearman correlations before averaging. In addition, the statistical comparisons in Appendix~\ref{app:statresults} are conducted on Fisher $z$-transformed correlations. The Fisher transform is,
\begin{equation}
z \;=\; \frac{1}{2}\ln\!\left(\frac{1+\rho}{1-\rho}\right),
\end{equation}
where $\rho$ denotes the Spearman correlation.
\begin{table}[htb!]
\centering
\small
\setlength{\tabcolsep}{5pt}
\begin{tabular}{l*{3}{c}}
\toprule
& \textbf{Llama-11B} & \textbf{Gemma-12B} & \textbf{Qwen-14B}\\
\midrule
$C_a$ (single)          & $-0.03{\sd{0.78}}$ & $1.27{\sd{2.08}}$ & $1.42{\sd{1.97}}$ \\
$C_a \mid (C_b = 0)$    & $0.28{\sd{1.18}}$ & $0.42{\sd{1.18}}$ & $1.62{\sd{2.10}}$\\
$C_a \mid (C_b = 1)$    & $-0.13{\sd{0.78}}$ & $0.03{\sd{0.72}}$ & $1.57{\sd{1.93}}$\\
$C_a \mid (C_b = 2)$    & $0.03{\sd{0.79}}$ & $-0.11{\sd{1.60}}$ & $1.21{\sd{1.61}}$\\
$C_a \mid (C_b = 3)$    & $0.10{\sd{0.76}}$ & $0.01{\sd{0.86}}$ & $1.36{\sd{2.01}}$\\
$C_a \mid (C_b = 4)$    & $-0.13{\sd{1.17}}$ & $-0.10{\sd{1.26}}$ & $1.05{\sd{1.63}}$\\
$C_a \mid C_b$ fixed    & $0.03{\sd{0.96}}$ & $0.05{\sd{1.17}}$ & $1.36{\sd{1.87}}$\\
$C_a \mid C_b$ rand     & $0.02{\sd{1.19}}$ & $0.34{\sd{1.48}}$ & $1.22{\sd{1.60}}$ \\
\midrule
$C_b$ (single)          & $2.45{\sd{2.92}}$ & $4.27{\sd{3.23}}$ & $3.83{\sd{3.18}}$\\
$C_b \mid (C_a = 0)$    & $1.12{\sd{1.37}}$ & $3.24{\sd{3.17}}$ & $3.27{\sd{3.04}}$\\
$C_b \mid (C_a = 1)$    & $2.33{\sd{2.33}}$ & $1.97{\sd{2.15}}$ & $2.86{\sd{2.95}}$\\
$C_b \mid (C_a = 2)$    & $1.50{\sd{1.73}}$ & $2.16{\sd{2.25}}$ & $2.81{\sd{2.86}}$\\
$C_b \mid (C_a = 3)$    & $1.54{\sd{1.72}}$ & $2.32{\sd{2.73}}$ & $3.29{\sd{3.04}}$\\
$C_b \mid (C_a = 4)$    & $2.60{\sd{2.73}}$ & $2.89{\sd{2.81}}$ & $3.13{\sd{2.91}}$\\
$C_b \mid C_a$ fixed    & $1.82{\sd{2.10}}$ & $2.51{\sd{2.68}}$ & $3.07{\sd{2.95}}$\\
$C_b \mid C_a$ rand     & $2.02{\sd{2.31}}$ & $1.96{\sd{2.29}}$ & $2.26{\sd{2.47}}$\\
\bottomrule
\end{tabular}
\caption{ARGUMENT generation \textbf{clarity--politeness} (Fisher-transformed Mean $\pm$ SD).}
\label{tab:arg-clarity-politeness-fisher-v2}
\end{table}

\begin{table}[htb!]
\centering
\small
\setlength{\tabcolsep}{5pt}
\begin{tabular}{l*{3}{c}}
\toprule
& \textbf{Llama-11B} & \textbf{Gemma-12B} & \textbf{Qwen-14B}\\
\midrule
$C_a$ (single)          & $3.38{\sd{2.97}}$ & $6.02{\sd{3.12}}$ & $6.74{\sd{2.79}}$ \\
$C_a \mid (C_b = 0)$    & $0.51{\sd{1.06}}$ & $5.29{\sd{3.28}}$ & $4.53{\sd{3.30}}$\\
$C_a \mid (C_b = 1)$    & $1.49{\sd{1.97}}$ & $5.70{\sd{3.25}}$ & $5.37{\sd{3.27}}$\\
$C_a \mid (C_b = 2)$    & $1.01{\sd{0.62}}$ & $4.35{\sd{3.16}}$ & $5.52{\sd{3.19}}$\\
$C_a \mid (C_b = 3)$    & $1.08{\sd{0.60}}$ & $3.73{\sd{2.95}}$ & $5.08{\sd{3.23}}$\\
$C_a \mid (C_b = 4)$    & $0.86{\sd{1.47}}$ & $3.92{\sd{3.02}}$ & $5.59{\sd{3.20}}$\\
$C_a \mid C_b$ fixed    & $0.99{\sd{1.29}}$ & $4.60{\sd{3.21}}$ & $5.22{\sd{3.24}}$\\
$C_a \mid C_b$ rand     & $0.54{\sd{0.67}}$ & $3.00{\sd{2.75}}$ & $3.91{\sd{3.13}}$ \\
\midrule
$C_b$ (single)          & $1.59{\sd{1.75}}$ & $6.19{\sd{3.07}}$ & $6.59{\sd{2.94}}$\\
$C_b \mid (C_a = 0)$    & $0.92{\sd{1.71}}$ & $6.37{\sd{3.09}}$ & $3.83{\sd{3.18}}$\\
$C_b \mid (C_a = 1)$    & $1.12{\sd{1.65}}$ & $3.11{\sd{2.91}}$ & $4.98{\sd{3.42}}$\\
$C_b \mid (C_a = 2)$    & $0.93{\sd{1.50}}$ & $2.64{\sd{2.57}}$ & $5.20{\sd{3.37}}$\\
$C_b \mid (C_a = 3)$    & $0.82{\sd{1.24}}$ & $2.38{\sd{2.42}}$ & $3.97{\sd{3.17}}$\\
$C_b \mid (C_a = 4)$    & $1.06{\sd{1.89}}$ & $2.50{\sd{2.64}}$ & $3.29{\sd{3.03}}$\\
$C_b \mid C_a$ fixed    & $0.97{\sd{1.61}}$ & $3.40{\sd{3.11}}$ & $4.25{\sd{3.30}}$\\
$C_b \mid C_a$ rand     & $0.92{\sd{1.68}}$ & $2.72{\sd{2.89}}$ & $3.32{\sd{3.00}}$\\
\bottomrule
\end{tabular}
\caption{ARGUMENT generation \textbf{formality--assertiveness} (Fisher-transformed Mean $\pm$ SD).}
\label{tab:arg-form-assert-fisher}
\end{table}

\begin{table}[htb!]
\centering
\small
\setlength{\tabcolsep}{5pt}
\begin{tabular}{l*{3}{c}}
\toprule
& \textbf{Llama-11B} & \textbf{Gemma-12B} & \textbf{Qwen-14B}\\
\midrule
$C_a$ (single)          & $1.94{\sd{2.32}}$ & $4.59{\sd{3.33}}$ & $4.02{\sd{3.33}}$ \\
$C_a \mid (C_b = 0)$    & $0.52{\sd{1.13}}$ & $3.23{\sd{3.07}}$ & $3.37{\sd{3.09}}$\\
$C_a \mid (C_b = 1)$    & $1.03{\sd{1.64}}$ & $2.22{\sd{2.35}}$ & $4.05{\sd{3.42}}$\\
$C_a \mid (C_b = 2)$    & $0.76{\sd{1.15}}$ & $2.41{\sd{2.42}}$ & $2.45{\sd{2.53}}$\\
$C_a \mid (C_b = 3)$    & $1.25{\sd{1.60}}$ & $2.69{\sd{2.67}}$ & $3.15{\sd{3.12}}$\\
$C_a \mid (C_b = 4)$    & $1.26{\sd{1.82}}$ & $3.29{\sd{2.92}}$ & $3.01{\sd{2.97}}$\\
$C_a \mid C_b$ fixed    & $0.96{\sd{1.51}}$ & $2.77{\sd{2.72}}$ & $3.20{\sd{3.07}}$\\
$C_a \mid C_b$ rand     & $0.76{\sd{0.61}}$ & $2.72{\sd{2.91}}$ & $3.71{\sd{3.37}}$ \\
\midrule
$C_b$ (single)          & $1.91{\sd{2.02}}$ & $6.31{\sd{3.09}}$ & $4.96{\sd{3.34}}$\\
$C_b \mid (C_a = 0)$    & $1.04{\sd{1.66}}$ & $4.99{\sd{3.32}}$ & $2.54{\sd{2.62}}$\\
$C_b \mid (C_a = 1)$    & $1.33{\sd{2.00}}$ & $1.90{\sd{1.97}}$ & $2.82{\sd{2.84}}$\\
$C_b \mid (C_a = 2)$    & $1.33{\sd{1.80}}$ & $2.01{\sd{2.43}}$ & $2.24{\sd{2.48}}$\\
$C_b \mid (C_a = 3)$    & $1.03{\sd{1.38}}$ & $2.02{\sd{2.28}}$ & $1.66{\sd{1.69}}$\\
$C_b \mid (C_a = 4)$    & $1.27{\sd{1.85}}$ & $1.58{\sd{1.71}}$ & $2.25{\sd{2.50}}$\\
$C_b \mid C_a$ fixed    & $1.20{\sd{1.75}}$ & $2.50{\sd{2.70}}$ & $2.30{\sd{2.48}}$\\
$C_b \mid C_a$ rand     & $1.10{\sd{1.84}}$ & $1.56{\sd{1.72}}$ & $2.01{\sd{2.27}}$\\
\bottomrule
\end{tabular}
\caption{ARGUMENT generation \textbf{humour--persuasiveness} (Fisher-transformed Mean $\pm$ SD).}
\label{tab:arg-humour-persuasiveness-fisher}
\end{table}

\begin{table}[htb!]
\centering
\small
\setlength{\tabcolsep}{5pt}
\begin{tabular}{l*{3}{c}}
\toprule
& \textbf{Llama-11B} & \textbf{Gemma-12B} & \textbf{Qwen-14B}\\
\midrule
$C_a$ (single)          & $0.91{\sd{1.72}}$ & $3.80{\sd{3.21}}$ & $2.61{\sd{2.59}}$ \\
$C_a \mid (C_b = 0)$    & $-0.13{\sd{0.68}}$ & $0.45{\sd{0.43}}$ & $1.23{\sd{1.59}}$\\
$C_a \mid (C_b = 1)$    & $-0.02{\sd{0.66}}$ & $0.50{\sd{0.72}}$ & $1.28{\sd{1.56}}$\\
$C_a \mid (C_b = 2)$    & $-0.00{\sd{0.84}}$ & $0.68{\sd{1.17}}$ & $2.05{\sd{2.41}}$\\
$C_a \mid (C_b = 3)$    & $0.05{\sd{0.76}}$ & $0.64{\sd{1.14}}$ & $1.54{\sd{1.70}}$\\
$C_a \mid (C_b = 4)$    & $0.06{\sd{0.80}}$ & $0.41{\sd{0.66}}$ & $1.77{\sd{2.04}}$\\
$C_a \mid C_b$ fixed    & $-0.01{\sd{0.75}}$ & $0.54{\sd{0.87}}$ & $1.57{\sd{1.90}}$\\
$C_a \mid C_b$ rand     & $-0.08{\sd{0.74}}$ & $0.58{\sd{1.46}}$ & $1.10{\sd{1.33}}$ \\
\midrule
$C_b$ (single)          & $2.95{\sd{3.02}}$ & $6.10{\sd{3.10}}$ & $5.34{\sd{3.32}}$\\
$C_b \mid (C_a = 0)$    & $1.12{\sd{1.65}}$ & $3.90{\sd{3.14}}$ & $5.00{\sd{3.39}}$\\
$C_b \mid (C_a = 1)$    & $1.68{\sd{1.89}}$ & $2.31{\sd{2.17}}$ & $5.04{\sd{3.45}}$\\
$C_b \mid (C_a = 2)$    & $1.47{\sd{1.76}}$ & $1.69{\sd{1.69}}$ & $3.84{\sd{3.17}}$\\
$C_b \mid (C_a = 3)$    & $1.33{\sd{1.33}}$ & $1.69{\sd{1.70}}$ & $3.87{\sd{3.15}}$\\
$C_b \mid (C_a = 4)$    & $1.44{\sd{1.30}}$ & $2.43{\sd{2.43}}$ & $4.38{\sd{3.33}}$\\
$C_b \mid C_a$ fixed    & $1.41{\sd{1.61}}$ & $2.41{\sd{2.42}}$ & $4.43{\sd{3.32}}$\\
$C_b \mid C_a$ rand     & $2.01{\sd{2.59}}$ & $1.46{\sd{1.75}}$ & $3.65{\sd{3.20}}$\\
\bottomrule
\end{tabular}
\caption{STORY generation \textbf{clarity--politeness} (Fisher-transformed Mean $\pm$ SD).}
\label{tab:story-clarity-politeness-fisher}
\end{table}

\begin{table}[htb!]
\centering
\small
\setlength{\tabcolsep}{5pt}
\begin{tabular}{l*{3}{c}}
\toprule
& \textbf{Llama-11B} & \textbf{Gemma-12B} & \textbf{Qwen-14B}\\
\midrule
$C_a$ (single)          & $4.14{\sd{3.33}}$ & $7.56{\sd{2.16}}$ & $6.30{\sd{3.09}}$ \\
$C_a \mid (C_b = 0)$    & $1.87{\sd{2.64}}$ & $6.18{\sd{3.08}}$ & $5.12{\sd{3.37}}$\\
$C_a \mid (C_b = 1)$    & $2.54{\sd{2.75}}$ & $6.14{\sd{3.13}}$ & $5.36{\sd{3.29}}$\\
$C_a \mid (C_b = 2)$    & $2.24{\sd{2.50}}$ & $4.45{\sd{3.27}}$ & $5.86{\sd{3.23}}$\\
$C_a \mid (C_b = 3)$    & $1.64{\sd{1.71}}$ & $4.13{\sd{3.16}}$ & $6.04{\sd{3.18}}$\\
$C_a \mid (C_b = 4)$    & $1.86{\sd{2.04}}$ & $4.28{\sd{3.23}}$ & $6.81{\sd{2.86}}$\\
$C_a \mid C_b$ fixed    & $2.03{\sd{2.37}}$ & $5.04{\sd{3.29}}$ & $5.84{\sd{3.23}}$\\
$C_a \mid C_b$ rand     & $2.10{\sd{2.54}}$ & $4.08{\sd{3.09}}$ & $5.20{\sd{3.37}}$ \\
\midrule
$C_b$ (single)          & $1.47{\sd{1.96}}$ & $5.90{\sd{3.19}}$ & $5.17{\sd{3.32}}$\\
$C_b \mid (C_a = 0)$    & $1.10{\sd{1.85}}$ & $3.25{\sd{2.95}}$ & $3.36{\sd{3.19}}$\\
$C_b \mid (C_a = 1)$    & $1.33{\sd{2.23}}$ & $4.38{\sd{3.33}}$ & $3.27{\sd{3.04}}$\\
$C_b \mid (C_a = 2)$    & $0.90{\sd{1.40}}$ & $4.63{\sd{3.38}}$ & $3.72{\sd{3.17}}$\\
$C_b \mid (C_a = 3)$    & $1.07{\sd{1.64}}$ & $3.86{\sd{3.25}}$ & $4.09{\sd{3.27}}$\\
$C_b \mid (C_a = 4)$    & $0.70{\sd{1.12}}$ & $3.39{\sd{3.08}}$ & $3.38{\sd{3.08}}$\\
$C_b \mid C_a$ fixed    & $1.02{\sd{1.69}}$ & $3.90{\sd{3.23}}$ & $3.56{\sd{3.15}}$\\
$C_b \mid C_a$ rand     & $0.87{\sd{1.69}}$ & $3.01{\sd{2.97}}$ & $2.87{\sd{2.82}}$\\
\bottomrule
\end{tabular}
\caption{STORY generation \textbf{formality--assertiveness} (Fisher-transformed Mean $\pm$ SD).}
\label{tab:story-formality-assertiveness-fisher}
\end{table}

\begin{table}[htb!]
\centering
\small
\setlength{\tabcolsep}{5pt}
\begin{tabular}{l*{3}{c}}
\toprule
& \textbf{Llama-11B} & \textbf{Gemma-12B} & \textbf{Qwen-14B}\\
\midrule
$C_a$ (single)          & $2.34{\sd{2.60}}$ & $4.37{\sd{3.33}}$ & $3.67{\sd{3.20}}$ \\
$C_a \mid (C_b = 0)$    & $0.34{\sd{0.63}}$ & $2.07{\sd{1.95}}$ & $4.14{\sd{3.34}}$\\
$C_a \mid (C_b = 1)$    & $0.92{\sd{1.70}}$ & $2.17{\sd{2.24}}$ & $3.06{\sd{2.84}}$\\
$C_a \mid (C_b = 2)$    & $0.73{\sd{1.41}}$ & $1.95{\sd{2.01}}$ & $3.13{\sd{3.01}}$\\
$C_a \mid (C_b = 3)$    & $0.49{\sd{1.15}}$ & $1.51{\sd{1.53}}$ & $2.70{\sd{2.67}}$\\
$C_a \mid (C_b = 4)$    & $0.70{\sd{0.74}}$ & $1.34{\sd{1.00}}$ & $3.22{\sd{3.08}}$\\
$C_a \mid C_b$ fixed    & $0.64{\sd{1.21}}$ & $1.81{\sd{1.82}}$ & $3.25{\sd{3.02}}$\\
$C_a \mid C_b$ rand     & $0.54{\sd{1.17}}$ & $1.69{\sd{2.07}}$ & $2.89{\sd{2.82}}$ \\
\midrule
$C_b$ (single)          & $1.57{\sd{1.51}}$ & $4.98{\sd{3.24}}$ & $3.76{\sd{3.14}}$\\
$C_b \mid (C_a = 0)$    & $0.69{\sd{0.66}}$ & $2.06{\sd{2.10}}$ & $2.85{\sd{2.83}}$\\
$C_b \mid (C_a = 1)$    & $1.19{\sd{1.59}}$ & $1.43{\sd{1.33}}$ & $2.18{\sd{2.07}}$\\
$C_b \mid (C_a = 2)$    & $1.16{\sd{1.57}}$ & $1.19{\sd{1.85}}$ & $2.35{\sd{2.46}}$\\
$C_b \mid (C_a = 3)$    & $0.91{\sd{1.05}}$ & $1.06{\sd{0.54}}$ & $1.80{\sd{1.84}}$\\
$C_b \mid (C_a = 4)$    & $1.12{\sd{1.42}}$ & $1.10{\sd{1.42}}$ & $2.19{\sd{2.37}}$\\
$C_b \mid C_a$ fixed    & $1.01{\sd{1.32}}$ & $1.37{\sd{1.58}}$ & $2.28{\sd{2.35}}$\\
$C_b \mid C_a$ rand     & $1.11{\sd{1.61}}$ & $1.43{\sd{1.77}}$ & $2.20{\sd{2.37}}$\\
\bottomrule
\end{tabular}
\caption{STORY generation \textbf{humour--persuasiveness} (Fisher-transformed Mean $\pm$ SD).}
\label{tab:story-humour-persuasiveness-fisher}
\end{table}

\begin{table}[htb!]
\centering
\small
\setlength{\tabcolsep}{5pt}
\begin{tabular}{l*{3}{c}}
\toprule
& \textbf{Llama-11B} & \textbf{Gemma-12B} & \textbf{Qwen-14B}\\
\midrule
$C_a$ (single)          & $0.52{\sd{1.56}}$ & $0.43{\sd{1.62}}$ & $1.23{\sd{1.77}}$ \\
$C_a \mid (C_b = 0)$    & $-0.01{\sd{0.53}}$ & $0.11{\sd{0.66}}$ & $0.44{\sd{0.61}}$\\
$C_a \mid (C_b = 1)$    & $-0.13{\sd{1.15}}$ & $-0.15{\sd{0.74}}$ & $0.55{\sd{0.57}}$\\
$C_a \mid (C_b = 2)$    & $-0.07{\sd{0.60}}$ & $-0.60{\sd{0.86}}$ & $0.82{\sd{1.40}}$\\
$C_a \mid (C_b = 3)$    & $0.09{\sd{0.61}}$ & $-0.60{\sd{1.18}}$ & $0.66{\sd{1.49}}$\\
$C_a \mid (C_b = 4)$    & $0.17{\sd{0.61}}$ & $-0.69{\sd{0.77}}$ & $0.89{\sd{1.67}}$\\
$C_a \mid C_b$ fixed    & $0.01{\sd{0.74}}$ & $-0.39{\sd{0.91}}$ & $0.67{\sd{1.24}}$\\
$C_a \mid C_b$ rand     & $0.10{\sd{0.49}}$ & $-0.28{\sd{0.70}}$ & $0.62{\sd{1.11}}$ \\
\midrule
$C_b$ (single)          & $1.76{\sd{2.22}}$ & $4.56{\sd{3.17}}$ & $3.42{\sd{2.95}}$\\
$C_b \mid (C_a = 0)$    & $0.56{\sd{0.69}}$ & $0.70{\sd{0.81}}$ & $1.15{\sd{1.63}}$\\
$C_b \mid (C_a = 1)$    & $1.05{\sd{1.18}}$ & $1.37{\sd{1.00}}$ & $1.80{\sd{1.87}}$\\
$C_b \mid (C_a = 2)$    & $0.83{\sd{1.11}}$ & $2.01{\sd{1.58}}$ & $2.63{\sd{2.59}}$\\
$C_b \mid (C_a = 3)$    & $0.52{\sd{0.65}}$ & $2.23{\sd{1.87}}$ & $2.87{\sd{2.83}}$\\
$C_b \mid (C_a = 4)$    & $0.58{\sd{0.61}}$ & $2.83{\sd{2.36}}$ & $1.92{\sd{2.00}}$\\
$C_b \mid C_a$ fixed    & $0.71{\sd{0.90}}$ & $1.83{\sd{1.78}}$ & $2.07{\sd{2.30}}$\\
$C_b \mid C_a$ rand     & $0.62{\sd{1.79}}$ & $1.16{\sd{1.37}}$ & $1.43{\sd{1.31}}$\\
\bottomrule
\end{tabular}
\caption{STRUCTURED generation \textbf{clarity--politeness} (Fisher-transformed Mean $\pm$ SD).}
\label{tab:structured-clarity-politeness-fisher}
\end{table}

\begin{table}[htb!]
\centering
\small
\setlength{\tabcolsep}{5pt}
\begin{tabular}{l*{3}{c}}
\toprule
& \textbf{Llama-11B} & \textbf{Gemma-12B} & \textbf{Qwen-14B}\\
\midrule
$C_a$ (single)          & $1.61{\sd{1.52}}$ & $3.08{\sd{2.61}}$ & $5.39{\sd{3.35}}$ \\
$C_a \mid (C_b = 0)$    & $0.27{\sd{0.78}}$ & $1.38{\sd{1.87}}$ & $1.79{\sd{2.04}}$\\
$C_a \mid (C_b = 1)$    & $0.59{\sd{0.64}}$ & $1.58{\sd{1.79}}$ & $3.10{\sd{3.03}}$\\
$C_a \mid (C_b = 2)$    & $0.78{\sd{1.13}}$ & $2.14{\sd{2.41}}$ & $4.22{\sd{3.37}}$\\
$C_a \mid (C_b = 3)$    & $1.06{\sd{1.66}}$ & $2.80{\sd{2.75}}$ & $3.69{\sd{3.28}}$\\
$C_a \mid (C_b = 4)$    & $1.03{\sd{1.65}}$ & $2.95{\sd{3.01}}$ & $2.50{\sd{2.48}}$\\
$C_a \mid C_b$ fixed    & $0.75{\sd{1.27}}$ & $2.17{\sd{2.48}}$ & $3.06{\sd{3.00}}$\\
$C_a \mid C_b$ rand     & $0.40{\sd{1.16}}$ & $1.93{\sd{2.48}}$ & $3.04{\sd{3.07}}$ \\
\midrule
$C_b$ (single)          & $1.40{\sd{1.80}}$ & $3.88{\sd{3.06}}$ & $2.36{\sd{2.31}}$\\
$C_b \mid (C_a = 0)$    & $0.45{\sd{0.81}}$ & $0.48{\sd{1.11}}$ & $1.20{\sd{1.61}}$\\
$C_b \mid (C_a = 1)$    & $0.93{\sd{1.47}}$ & $1.27{\sd{1.04}}$ & $2.53{\sd{2.91}}$\\
$C_b \mid (C_a = 2)$    & $0.92{\sd{1.68}}$ & $2.06{\sd{2.11}}$ & $2.00{\sd{2.46}}$\\
$C_b \mid (C_a = 3)$    & $0.75{\sd{0.90}}$ & $2.58{\sd{2.51}}$ & $2.42{\sd{2.69}}$\\
$C_b \mid (C_a = 4)$    & $1.08{\sd{1.92}}$ & $1.95{\sd{1.82}}$ & $2.50{\sd{2.52}}$\\
$C_b \mid C_a$ fixed    & $0.83{\sd{1.43}}$ & $1.67{\sd{1.94}}$ & $2.13{\sd{2.51}}$\\
$C_b \mid C_a$ rand     & $0.69{\sd{1.21}}$ & $1.32{\sd{1.34}}$ & $2.16{\sd{2.66}}$\\
\bottomrule
\end{tabular}
\caption{STRUCTURED generation \textbf{formality--assertiveness} (Fisher-transformed Mean $\pm$ SD).}
\label{tab:structured-formality-assertiveness-fisher}
\end{table}

\begin{table}[htb!]
\centering
\small
\setlength{\tabcolsep}{5pt}
\begin{tabular}{l*{3}{c}}
\toprule
& \textbf{Llama-11B} & \textbf{Gemma-12B} & \textbf{Qwen-14B}\\
\midrule
$C_a$ (single)          & $1.27{\sd{1.29}}$ & $5.17{\sd{3.41}}$ & $3.36{\sd{3.10}}$ \\
$C_a \mid (C_b = 0)$    & $0.38{\sd{1.17}}$ & $3.13{\sd{3.01}}$ & $3.44{\sd{3.35}}$\\
$C_a \mid (C_b = 1)$    & $0.37{\sd{0.70}}$ & $3.04{\sd{2.96}}$ & $3.30{\sd{3.24}}$\\
$C_a \mid (C_b = 2)$    & $0.37{\sd{1.28}}$ & $2.75{\sd{2.88}}$ & $2.25{\sd{2.62}}$\\
$C_a \mid (C_b = 3)$    & $0.80{\sd{1.71}}$ & $3.01{\sd{2.97}}$ & $2.92{\sd{2.91}}$\\
$C_a \mid (C_b = 4)$    & $0.74{\sd{1.18}}$ & $3.74{\sd{3.25}}$ & $2.98{\sd{3.11}}$\\
$C_a \mid C_b$ fixed    & $0.53{\sd{1.26}}$ & $3.14{\sd{3.02}}$ & $2.98{\sd{3.07}}$\\
$C_a \mid C_b$ rand     & $0.24{\sd{0.66}}$ & $2.42{\sd{2.81}}$ & $3.01{\sd{3.20}}$ \\
\midrule
$C_b$ (single)          & $2.49{\sd{2.38}}$ & $6.81{\sd{2.76}}$ & $7.63{\sd{2.13}}$\\
$C_b \mid (C_a = 0)$    & $0.63{\sd{1.12}}$ & $4.55{\sd{3.28}}$ & $5.78{\sd{3.34}}$\\
$C_b \mid (C_a = 1)$    & $1.47{\sd{2.16}}$ & $2.90{\sd{2.80}}$ & $3.20{\sd{2.98}}$\\
$C_b \mid (C_a = 2)$    & $1.34{\sd{1.82}}$ & $3.22{\sd{3.07}}$ & $3.63{\sd{3.33}}$\\
$C_b \mid (C_a = 3)$    & $1.17{\sd{1.64}}$ & $3.46{\sd{3.14}}$ & $3.10{\sd{3.03}}$\\
$C_b \mid (C_a = 4)$    & $1.31{\sd{1.84}}$ & $3.73{\sd{3.25}}$ & $3.35{\sd{3.20}}$\\
$C_b \mid C_a$ fixed    & $1.18{\sd{1.76}}$ & $3.57{\sd{3.15}}$ & $3.81{\sd{3.32}}$\\
$C_b \mid C_a$ rand     & $1.10{\sd{1.85}}$ & $1.70{\sd{1.88}}$ & $2.40{\sd{2.68}}$\\
\bottomrule
\end{tabular}
\caption{STRUCTURED generation \textbf{humour--persuasiveness} (Fisher-transformed Mean $\pm$ SD).}
\label{tab:structured-humour-persuasiveness-fisher}
\end{table}

\onecolumn
\newpage
\section{Statistical Test Results}
\label{app:statresults}

We conduct paired $t$-tests on Fisher $z$-transformed Spearman correlations for each medium-sized model and concept pair in the argument-generation dataset. Because Spearman correlations are bounded and skewed near 1, we first apply the Fisher transformation. For each sample, we compute the per-sample difference between the single- and dual-concept Fisher values and perform a one-sided paired $t$-test. 

\medskip
\noindent The null hypothesis is that the mean difference is zero, corresponding to no change in performance, while the alternative hypothesis is that single-concept control is stronger than dual-concept control. The resulting $t$-statistics and one-sided $p$-values are reported below. Large positive $t$-values accompanied by extremely small $p$-values indicate that introducing a secondary concept significantly harms controllability.

\subsection{Random Secondary Results}
\begin{table}[htb!]
\centering
\small
\begin{tabular}{llc cc cc cc}
\toprule
\textbf{Primary} & \textbf{Secondary} &
& \multicolumn{2}{c}{\textbf{Llama-11B}} 
& \multicolumn{2}{c}{\textbf{Gemma-12B}} 
& \multicolumn{2}{c}{\textbf{Qwen-14B}} \\
\cmidrule(lr){4-5} \cmidrule(lr){6-7} \cmidrule(lr){8-9}
& &
& $t$ & $p$ (one-sided)
& $t$ & $p$ (one-sided)
& $t$ & $p$ (one-sided) \\
\midrule
assertiveness & formality &
& 2.37 & $1.01\times10^{-2}$
& 6.98 & $5.30\times10^{-10}$
& 6.79 & $1.22\times10^{-9}$ \\

formality & assertiveness &
& 8.34 & $1.45\times10^{-12}$
& 6.13 & $1.99\times10^{-8}$
& 5.71 & $1.10\times10^{-7}$ \\

clarity & politeness &
& -0.31 & $6.22\times10^{-1}$
& 3.27 & $8.06\times10^{-4}$
& 0.66 & $2.56\times10^{-1}$ \\

politeness & clarity &
& 1.18 & $1.21\times10^{-1}$
& 5.31 & $5.48\times10^{-7}$
& 3.44 & $4.77\times10^{-4}$ \\

humor & persuasiveness &
& 4.35 & $2.14\times10^{-5}$
& 3.63 & $2.61\times10^{-4}$
& 0.53 & $2.99\times10^{-1}$ \\

persuasiveness & humor &
& 2.43 & $8.80\times10^{-3}$
& 12.48 & $3.45\times10^{-20}$
& 6.33 & $8.58\times10^{-9}$ \\
\bottomrule
\end{tabular}
\caption{ARGUMENT generation paired one-sided $t$-tests on Fisher $z$-transformed Spearman correlations comparing single-concept and dual-concept with \textbf{random} secondary control across medium-sized models.}
\label{tab:paired_ttests_argument_generation}
\end{table}

\begin{table}[htb!]
\centering
\small
\begin{tabular}{llc cc cc cc}
\toprule
\textbf{Primary} & \textbf{Secondary} &
& \multicolumn{2}{c}{\textbf{Llama-11B}} 
& \multicolumn{2}{c}{\textbf{Gemma-12B}} 
& \multicolumn{2}{c}{\textbf{Qwen-14B}} \\
\cmidrule(lr){4-5} \cmidrule(lr){6-7} \cmidrule(lr){8-9}
& &
& $t$ & $p$ (one-sided)
& $t$ & $p$ (one-sided)
& $t$ & $p$ (one-sided) \\
\midrule
assertiveness & formality &
& 1.99 & $2.50\times10^{-2}$
& 5.62 & $1.59\times10^{-7}$
& 4.46 & $1.41\times10^{-5}$ \\

formality & assertiveness &
& 4.14 & $4.60\times10^{-5}$
& 8.45 & $9.09\times10^{-13}$
& 1.91 & $3.03\times10^{-2}$ \\

clarity & politeness &
& 4.23 & $3.35\times10^{-5}$
& 7.41 & $8.31\times10^{-11}$
& 4.91 & $2.92\times10^{-6}$ \\

politeness & clarity &
& 2.10 & $1.96\times10^{-2}$
& 11.49 & $2.01\times10^{-18}$
& 3.22 & $9.65\times10^{-4}$ \\

humor & persuasiveness &
& 5.63 & $1.51\times10^{-7}$
& 6.27 & $1.10\times10^{-8}$
& 1.74 & $4.30\times10^{-2}$ \\

persuasiveness & humor &
& 1.65 & $5.12\times10^{-2}$
& 8.36 & $1.37\times10^{-12}$
& 3.44 & $4.83\times10^{-4}$ \\
\bottomrule
\end{tabular}
\caption{STORY generation paired one-sided $t$-tests on Fisher $z$-transformed Spearman correlations comparing single-concept and dual-concept with \textbf{random} secondary control across medium-sized models.}
\label{tab:paired_ttests_story}
\end{table}

\begin{table}[htb!]
\centering
\small
\begin{tabular}{llc cc cc cc}
\toprule
\textbf{Primary} & \textbf{Secondary} &
& \multicolumn{2}{c}{\textbf{Llama-11B}} 
& \multicolumn{2}{c}{\textbf{Gemma-12B}} 
& \multicolumn{2}{c}{\textbf{Qwen-14B}} \\
\cmidrule(lr){4-5} \cmidrule(lr){6-7} \cmidrule(lr){8-9}
& &
& $t$ & $p$ (one-sided)
& $t$ & $p$ (one-sided)
& $t$ & $p$ (one-sided) \\
\midrule
assertiveness & formality &
& 3.92 & $9.69\times10^{-5}$
& 7.14 & $2.71\times10^{-10}$
& 0.47 & $3.20\times10^{-1}$ \\

formality & assertiveness &
& 5.76 & $8.94\times10^{-8}$
& 3.44 & $4.73\times10^{-4}$
& 5.04 & $1.60\times10^{-6}$ \\

clarity & politeness &
& 2.40 & $9.41\times10^{-3}$
& 3.37 & $6.04\times10^{-4}$
& 3.18 & $1.07\times10^{-3}$ \\

politeness & clarity &
& 3.40 & $5.49\times10^{-4}$
& 9.42 & $1.33\times10^{-14}$
& 5.28 & $6.29\times10^{-7}$ \\

humor & persuasiveness &
& 5.66 & $1.35\times10^{-7}$
& 5.44 & $3.32\times10^{-7}$
& 0.78 & $2.20\times10^{-1}$ \\

persuasiveness & humor &
& 4.22 & $3.42\times10^{-5}$
& 13.97 & $9.52\times10^{-23}$
& 12.99 & $4.49\times10^{-21}$ \\
\bottomrule
\end{tabular}
\caption{STRUCTURED generation paired one-sided $t$-tests on Fisher $z$-transformed Spearman correlations comparing single-concept and dual-concept with \textbf{random} secondary control across medium-sized models.}
\label{tab:paired_ttests_dart}
\end{table}

\subsection{Constant Secondary Results}
For the constant secondary statistical tests, we average the fisher scores across the five fixed concept levels before performing the test.
\begin{table}[htb!]
\centering
\small
\begin{tabular}{llc cc cc cc}
\toprule
\textbf{Primary} & \textbf{Secondary} &
& \multicolumn{2}{c}{\textbf{Llama-11B}} 
& \multicolumn{2}{c}{\textbf{Gemma-12B}} 
& \multicolumn{2}{c}{\textbf{Qwen-14B}} \\
\cmidrule(lr){4-5} \cmidrule(lr){6-7} \cmidrule(lr){8-9}
& &
& $t$ & $p$ (one-sided)
& $t$ & $p$ (one-sided)
& $t$ & $p$ (one-sided) \\
\midrule
assertiveness & formality &
& 2.75 & $3.76\times10^{-3}$
& 7.22 & $1.90\times10^{-10}$
& 6.84 & $9.85\times10^{-10}$ \\

formality & assertiveness &
& 6.78 & $1.28\times10^{-9}$
& 3.59 & $2.96\times10^{-4}$
& 4.14 & $4.59\times10^{-5}$ \\

clarity & politeness &
& -0.53 & $7.02\times10^{-1}$
& 5.01 & $1.80\times10^{-6}$
& 0.24 & $4.07\times10^{-1}$ \\

politeness & clarity &
& 1.74 & $4.28\times10^{-2}$
& 4.47 & $1.40\times10^{-5}$
& 1.82 & $3.61\times10^{-2}$ \\

humor & persuasiveness &
& 3.60 & $2.89\times10^{-4}$
& 4.81 & $3.87\times10^{-6}$
& 1.87 & $3.25\times10^{-2}$ \\

persuasiveness & humor &
& 2.73 & $3.92\times10^{-3}$
& 10.07 & $8.01\times10^{-16}$
& 6.49 & $4.39\times10^{-9}$ \\
\bottomrule
\end{tabular}
\caption{ARGUMENT generation paired one-sided $t$-tests on Fisher $z$-transformed Spearman correlations comparing single-concept and dual-concept with \textbf{constant} secondary control across medium-sized models.}
\label{tab:paired_ttests_sec_const_arg}
\end{table}

\begin{table}[htb!]
\centering
\small
\begin{tabular}{llc cc cc cc}
\toprule
\textbf{Primary} & \textbf{Secondary} &
& \multicolumn{2}{c}{\textbf{Llama-11B}} 
& \multicolumn{2}{c}{\textbf{Gemma-12B}} 
& \multicolumn{2}{c}{\textbf{Qwen-14B}} \\
\cmidrule(lr){4-5} \cmidrule(lr){6-7} \cmidrule(lr){8-9}
& &
& $t$ & $p$ (one-sided)
& $t$ & $p$ (one-sided)
& $t$ & $p$ (one-sided) \\
\midrule
assertiveness & formality &
& 2.07 & $2.12\times10^{-2}$
& 4.96 & $2.22\times10^{-6}$
& 3.86 & $1.21\times10^{-4}$ \\

formality & assertiveness &
& 5.14 & $1.07\times10^{-6}$
& 8.66 & $3.68\times10^{-13}$
& 1.18 & $1.21\times10^{-1}$ \\

clarity & politeness &
& 4.49 & $1.30\times10^{-5}$
& 8.74 & $2.52\times10^{-13}$
& 3.22 & $9.42\times10^{-4}$ \\

politeness & clarity &
& 4.49 & $1.27\times10^{-5}$
& 9.53 & $8.49\times10^{-15}$
& 2.19 & $1.59\times10^{-2}$ \\

humor & persuasiveness &
& 5.50 & $2.61\times10^{-7}$
& 6.12 & $2.00\times10^{-8}$
& 1.06 & $1.47\times10^{-1}$ \\

persuasiveness & humor &
& 2.78 & $3.44\times10^{-3}$
& 9.13 & $4.76\times10^{-14}$
& 3.80 & $1.45\times10^{-4}$ \\
\bottomrule
\end{tabular}
\caption{STORY generation paired one-sided $t$-tests on Fisher $z$-transformed Spearman correlations comparing single-concept and dual-concept with \textbf{constant} secondary control across medium-sized models.}
\label{tab:paired_ttests_sec_const_story}
\end{table}

\begin{table}[htb!]
\centering
\small
\begin{tabular}{llc cc cc cc}
\toprule
\textbf{Primary} & \textbf{Secondary} &
& \multicolumn{2}{c}{\textbf{Llama-11B}} 
& \multicolumn{2}{c}{\textbf{Gemma-12B}} 
& \multicolumn{2}{c}{\textbf{Qwen-14B}} \\
\cmidrule(lr){4-5} \cmidrule(lr){6-7} \cmidrule(lr){8-9}
& &
& $t$ & $p$ (one-sided)
& $t$ & $p$ (one-sided)
& $t$ & $p$ (one-sided) \\
\midrule
assertiveness & formality &
& 3.09 & $1.41\times10^{-3}$
& 6.21 & $1.43\times10^{-8}$
& 0.78 & $2.19\times10^{-1}$ \\

formality & assertiveness &
& 4.76 & $5.00\times10^{-6}$
& 3.00 & $1.85\times10^{-3}$
& 6.03 & $2.95\times10^{-8}$ \\

clarity & politeness &
& 2.73 & $3.95\times10^{-3}$
& 4.22 & $3.40\times10^{-5}$
& 2.74 & $3.88\times10^{-3}$ \\

politeness & clarity &
& 4.15 & $4.40\times10^{-5}$
& 7.44 & $7.52\times10^{-11}$
& 4.42 & $1.67\times10^{-5}$ \\

humor & persuasiveness &
& 4.49 & $1.30\times10^{-5}$
& 4.78 & $4.38\times10^{-6}$
& 1.10 & $1.37\times10^{-1}$ \\

persuasiveness & humor &
& 4.49 & $1.30\times10^{-5}$
& 10.51 & $1.24\times10^{-16}$
& 12.64 & $1.85\times10^{-20}$ \\
\bottomrule
\end{tabular}
\caption{STRUCTURED generation paired one-sided $t$-tests on Fisher $z$-transformed Spearman correlations comparing single-concept and dual-concept with \textbf{constant} secondary control across medium-sized models.}
\label{tab:paired_ttests_sec_const_dart}
\end{table}

\onecolumn
\newpage
\section{Human Evaluation}
\label{app:humaneval}

We use human evaluation to validate the performance of LLM-as-judge. We provided three humans with the LLM judge's ranking for 5 levels of single-concept generation e.g. humor. We asked them to state whether they agree, somewhat agree, or disagree with the LLM's ranking. These were given scores of 1, 0.5 and 0 respectively. Each human was given 6 examples corresponding to the 6 individual concepts that we explore. The results are shown in the table below:

\begin{table}[htb!]
\centering
\small
\begin{tabular}{lcccc}
\toprule
 & \textbf{Human1} & \textbf{Human2} & \textbf{Human3} & \textbf{Overall} \\
\midrule
Score & 4/6 & 5/6 & 5/6 & 14/18 \\
\bottomrule
\end{tabular}
\caption{Human evaluation results}
\label{tab:human_scores}
\end{table}

\onecolumn
\newpage
\section{Pairwise vs Listwise LLM Judge}
\label{app:pairvslist}
We had run preliminary experiments asking the judge-LLM to perform a single inference ranking of all responses (where responses are provided in a randomized order). We observed strong position bias: the first-presented sample was disproportionately ranked lowest. The table below shows the fraction of cases where the first item was ranked last (the first in the output of the list from the LLM) (see Table~\ref{tab:model-preference-fractions}).
\begin{table}[htb!]
\centering
\small
\begin{tabular}{lccc}
\toprule
\textbf{Setting} & \textbf{Total Samples} & \textbf{Llama70b Fraction} & \textbf{Qwen72b Fraction} \\
\midrule
Humor (single) & 75 & 0.387 & 0.667 \\
Persuasiveness (single) & 75 & 0.160 & 0.373 \\
Humor $\mid$ Persuasiveness (random) & 75 & 0.133 & 0.040 \\
Persuasiveness $\mid$ Humor (random) & 75 & 0.160 & 0.107 \\
Humor $\mid$ Persuasiveness (constant) & 375 & 0.389 & 0.725 \\
Persuasiveness $\mid$ Humor (constant) & 375 & 0.205 & 0.483 \\
\bottomrule
\end{tabular}
\caption{Model preference fractions across different control settings}
\label{tab:model-preference-fractions}
\end{table}

\onecolumn
\newpage
\section{Score Tie Proportions}
\label{app:scoreties}

We visualize the score ties per concept per model per dataset. For each concept, we calculate the proportion of ties over the single-concept, dual-concept random, and dual-concept constant experiments.

\begin{figure}[htb!]
\centering
\includegraphics[width=0.32\textwidth]{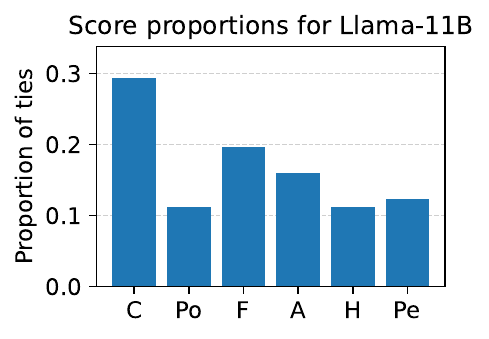}
\includegraphics[width=0.32\textwidth]{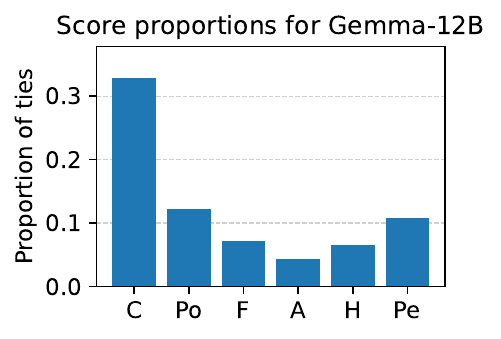}
\includegraphics[width=0.32\textwidth]{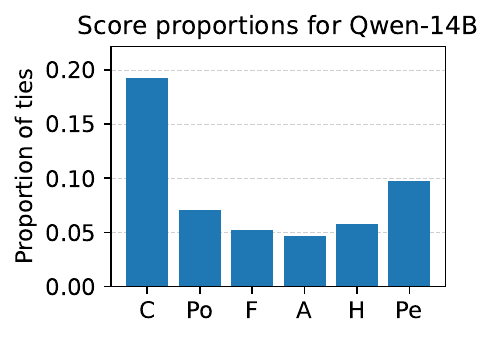}
\caption{ARGUMENT generation - score tie proportions across six concepts for each model.}
\label{fig:score_proportions_models_arg}
\end{figure}

\begin{figure}[htb!]
\centering
\includegraphics[width=0.32\textwidth]{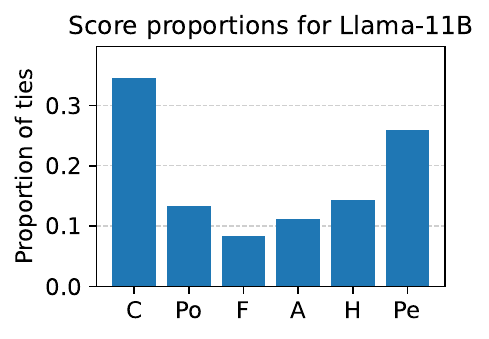}
\includegraphics[width=0.32\textwidth]{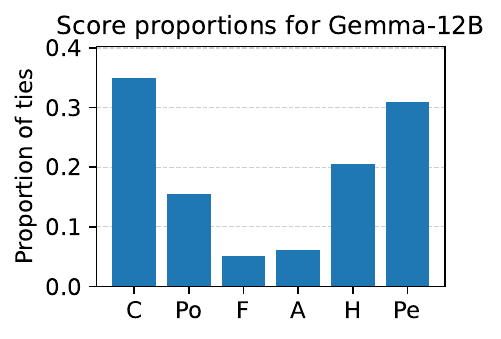}
\includegraphics[width=0.32\textwidth]{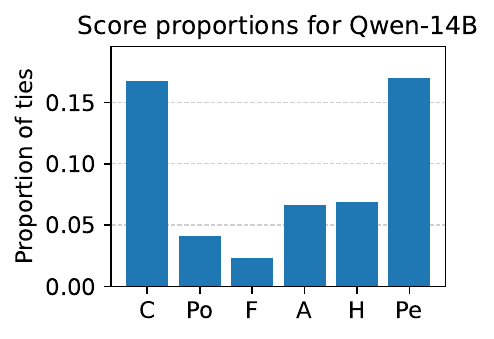}
\caption{STORY generation - score tie proportions across six concepts for each model.}
\label{fig:score_proportions_models_story}
\end{figure}

\begin{figure}[htb!]
\centering
\includegraphics[width=0.32\textwidth]{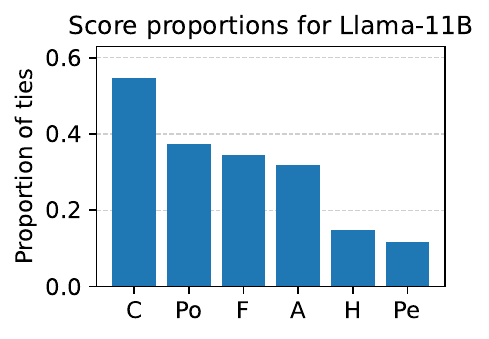}
\includegraphics[width=0.32\textwidth]{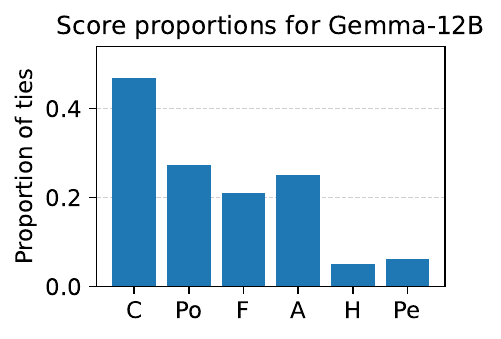}
\includegraphics[width=0.32\textwidth]{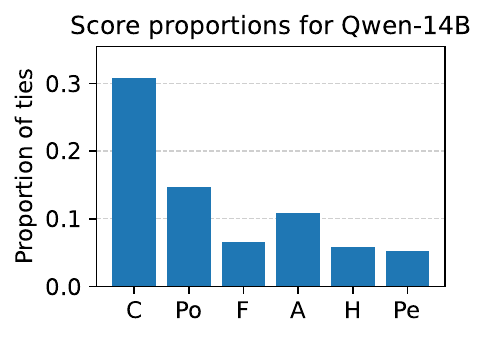}
\caption{STRUCTURED generation - score tie proportions across six concepts for each model.}
\label{fig:score_proportions_models_story}
\end{figure}
\newpage
\section{LLM Prompts}
\label{app:prompts}
\begin{mdframed}[backgroundcolor=gray!10, linewidth=1pt, outermargin=0cm]
\ttfamily
You are given some text: "I was driving around a neighborhood. I saw my friend outside his house. He had a large hose. Water was coming out if it. He was draining his pool."

Task Description: Your task is to write a short story that continues on from the text provided, and incorporates the concept given below at the desired level.

Concept: {concept}

Desired Concept Level: 1/4

Level 0/4 implies no presence of {concept}, and level 4/4 implies maximal presence of {concept}.

Your output should only be the story, without any additional text or explanation.
\end{mdframed}
\captionof{figure}{Single Concept Prompt Example. Task: Story Generation.}

\begin{mdframed}[backgroundcolor=gray!10, linewidth=1pt, outermargin=0cm]
\ttfamily
You are given some text: "I was driving around a neighborhood. I saw my friend outside his house. He had a large hose. Water was coming out if it. He was draining his pool."

Task Description: Your task is to write a short story that continues on from the text provided, and incorporates the concepts given below at the desired levels.

Concepts: {Concept A}, {Concept B}

Desired Concept Levels: {Concept A} at level 2/4 and {Concept B} at level 4/4

Level 0/4 implies no presence of the concept, and level 4/4 implies maximal presence of the concept.

Your output should only be the story, without any additional text or explanation.
\end{mdframed}
\captionof{figure}{Multi Concept Prompt Example. Task: Story Generation}

\begin{mdframed}[backgroundcolor=gray!10, linewidth=1pt, outermargin=0cm]
\ttfamily
You are given a claim: "Social media should not be required to verify user identities"

Task Description: Your task is to write a brief argument supporting this claim that incorporates the concept given below at the desired level.

Concept: {Concept A}

Desired Concept Level: 1/4

Level 0/4 implies no presence of {Concept A}, and level 4/4 implies maximal presence of {Concept A}.

Your output should only be the argument, without any additional text or explanation.
\end{mdframed}
\captionof{figure}{Single Concept Prompt Example. Task: Argument Generation.}

\begin{mdframed}[backgroundcolor=gray!10, linewidth=1pt, outermargin=0cm]
\ttfamily
You are given a claim: "Social media should not be required to verify user identities"

Your task is to write a brief argument supporting this claim that incorporates the concepts given below at the desired levels.

Concepts: {Concept A}, {Concept B}

Desired Concept Levels: {Concept A} at level 2/4 and {Concept B} at level 4/4

Level 0/4 implies no presence of the concept, and level 4/4 implies maximal presence of the concept.

Your output should only be the argument, without any additional text or explanation.
\end{mdframed}
\captionof{figure}{Multi Concept Prompt Example. Task: Argument Generation}

\begin{mdframed}[backgroundcolor=gray!10, linewidth=1pt, outermargin=0cm]
\ttfamily
You are given the following structured data: ["Sundiata Gaines", "TEAM", "Georgia"]

Task Description: Your task is to write a textual description from the provided structured data and NOTHING ELSE. The context may not be clear or limited, but it is your job to infer the context and you must provide a response. Your response should incorporate the concept given below at the desired level.

Concept: {Concept A}

Desired Concept Level: 2/4

Level 0/4 implies no presence of {Concept A}, and level 4/4 implies maximal presence of {Concept A}.

Your output MUST ONLY be the textual description, without any additional text or explanation.
\end{mdframed}
\captionof{figure}{Single Concept Prompt Example. Task: Structured Data Generation.}

\begin{mdframed}[backgroundcolor=gray!10, linewidth=1pt, outermargin=0cm]
\ttfamily
You are given the following structured data: ["Sundiata Gaines", "TEAM", "Georgia"]

Task Description: Your task is to write a textual description from the provided structured data and NOTHING ELSE. The context may not be clear or limited, but it is your job to infer the context and you must provide a response. Your response should incorporate the concepts given below at the desired levels.

Concepts: {Concept A}, {Concept B}

Desired Concept Levels: {Concept A} at level 1/4 and {Concept B} at level 3/4

Level 0/4 implies no presence of the concept, and level 4/4 implies maximal presence of the concept.

Your output MUST ONLY be the textual description, without any additional text or explanation.
\end{mdframed}
\captionof{figure}{Multi Concept Prompt Example. Task: Structured Data Generation}

\begin{mdframed}[backgroundcolor=gray!10, linewidth=1pt, outermargin=0cm]
\ttfamily
Which of these two statements shows a greater level of ‘formality’?

A: """
Social media platforms are meant to be casual and relaxed spaces for people to connect and share their thoughts. Requiring verification of user identities would make these spaces feel more formal and restrictive, which would go against the very nature of social media. It would also create unnecessary barriers for people who want to engage online, especially those who may not have the necessary documentation or resources to verify their identities. By keeping social media informal, we can maintain its accessibility and openness, allowing people to express themselves freely and connect with others in a more organic way.
"""

B: """
Requiring social media platforms to verify user identities would undermine the informal nature of online interactions, which is a fundamental aspect of their appeal. The internet's ability to facilitate anonymous and pseudonymous communication has been a driving force behind its widespread adoption and the free exchange of ideas it enables. Formal identity verification would introduce a level of formality that could stifle online discourse, making it less accessible and less inclusive for those who value the anonymity that the internet provides.
"""

Think step by step if needed, but keep it brief. Then provide your final choice clearly in the following format:
<Answer>A</Answer> or <Answer>B</Answer>
Do not leave the answer blank.
\end{mdframed}
\captionof{figure}{LLM-as-a-Judge Prompt Example}
\newpage
\section{Spearman Rank Distributions} \label{sec:app-hist}
\begin{figure}[htbp]
\centering

\begin{subfigure}{0.45\textwidth}
    \centering
    \includegraphics[width=\linewidth]{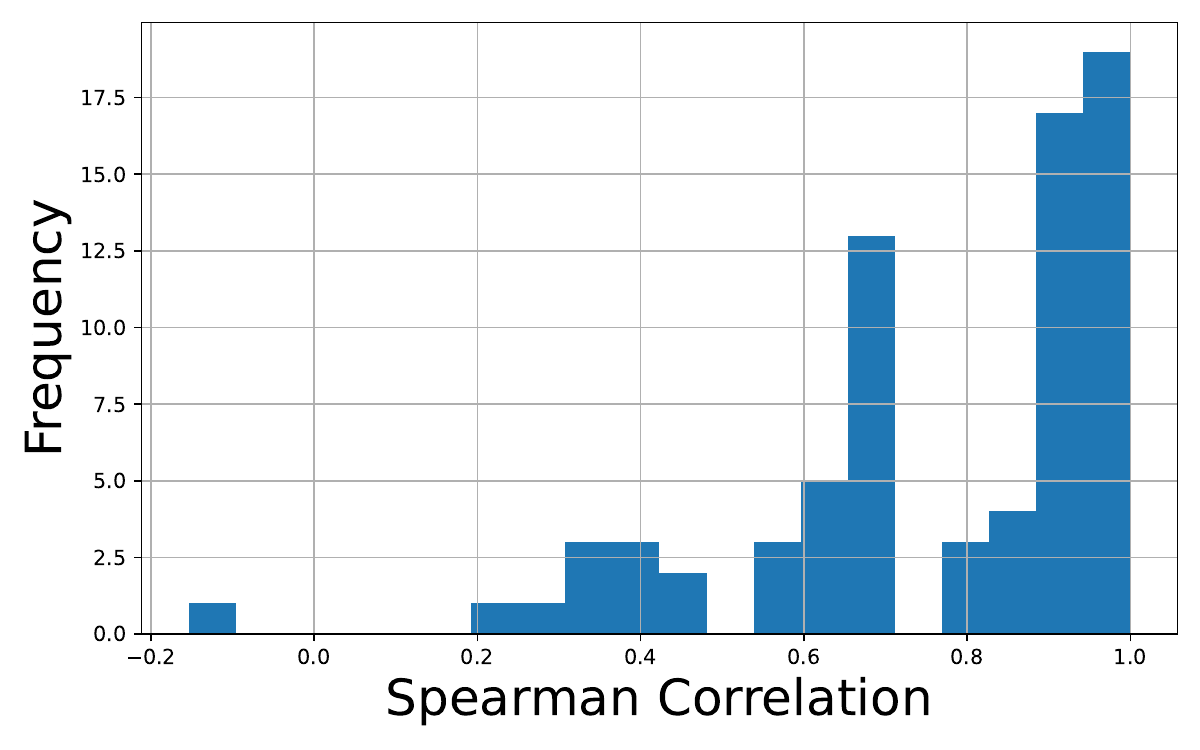}
    \caption{Humor. Model: Llama}
    \label{fig:1a}
\end{subfigure}
\hfill
\begin{subfigure}{0.45\textwidth}
    \centering
    \includegraphics[width=\linewidth]{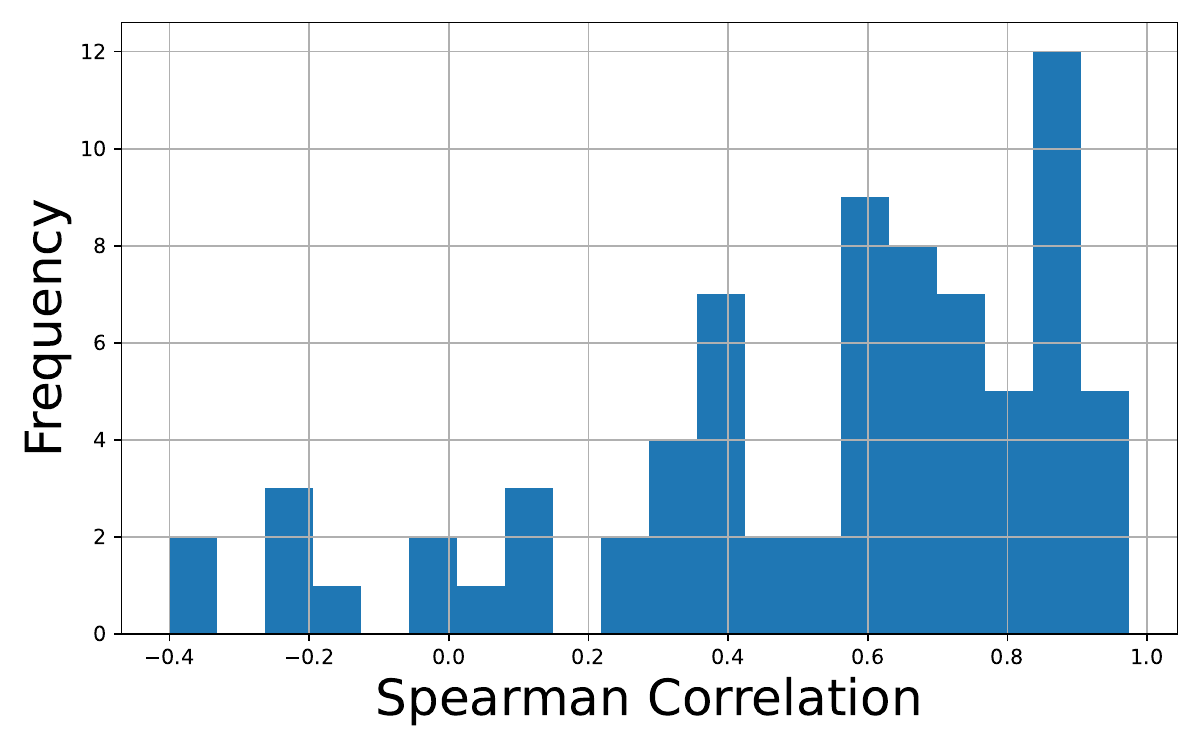}
    \caption{Humor | Persuasiveness. Model: Llama}
    \label{fig:1b}
\end{subfigure}





\caption{Example of the distribution of Spearman correlation values between the desired levels and rank of the target concept. Result shown for argument generation task, using Llama 3-11B, with the target concept only and secondary concept level \textit{randomly} sampled. For example, ``Humor $|$ Persuasiveness'' denotes responses generated independently for each humor level (target concept) while persuasiveness is randomly set for each inference.}
\label{fig:main}
\end{figure}
\newpage
\section{Example Outputs}
\label{app:exampleoutputs}
\begin{table*}[h!]
\centering
\begin{tabularx}{\textwidth}{|l|X|}
\hline
\multicolumn{2}{|c|}{\textbf{Claim:} Governments and technology companies must do more to protect online privacy and security.} \\
\hline
\textbf{Level 0} & The increasing reliance on technology has led to a significant shift in the way we live, work, and interact with one another. However, this shift has also raised concerns about online privacy and security. Governments and technology companies have a responsibility to protect their citizens' and users' personal data from falling into the wrong hands. This is not just a moral obligation, but also a necessary step to prevent the misuse of sensitive information. By taking proactive measures to safeguard online privacy and security, governments and technology companies can help build trust and ensure that the benefits of technology are accessible to all. \\
\hline
\textbf{Level 1} & We're living in a world where our online activities are being tracked, monitored, and sold to the highest bidder. It's like we're all wearing a neon sign that screams \"Hey, I'm a potential customer, come and get me!\" Governments and tech companies are like the bouncers at a nightclub, collecting our personal data and serving it up to advertisers like a never-ending happy hour special. But let's be real, folks, our online security is a joke. It's like trying to keep a secret in a kindergarten classroom – it's just not happening. We need governments and tech companies to step up their game and give us some real protection. \\
\hline
\textbf{Level 2} & We're living in a world where our online activities are being tracked, monitored, and sold to the highest bidder. It's like we're all wearing a neon sign that screams "Hey, I'm a potential customer, come and get me!" Governments and tech companies are like the bumbling security guards who can't even keep the coffee machine from being hacked, let alone our personal data. It's time for them to step up their game and give us some real protection. After all, we're not just talking about our online security, we're talking about our digital dignity. \\
\hline
\textbf{Level 3} & The age-old adage: "if you're not paying for the product, you are the product.\" It's a clever way of saying that when we use free online services, we're essentially trading our personal data for the privilege of using them. But let's be real, who doesn't love a good bargain? The problem is, our personal data is worth a lot more than a few dollars off our next purchase. It's like selling our soul to the digital devil, and then wondering why our online lives are being hijacked by bots and scammers. Governments and tech companies need to step up their game and protect our online privacy and security. After all, we're not just talking about our data – we're talking about our digital dignity. \\
\hline
\textbf{Level 4} & The cat's out of the bag – or should I say, the cat's out of the cloud? Governments and tech companies are supposed to be the guardians of our online security, but it seems they're more like the bumbling superheroes of the digital world. They're always one step behind the bad guys, tripping over their own wires and leaving our personal info exposed for the world to see. It's like they're playing a game of digital whack-a-mole, where every time they squash one security threat, another one pops up in its place. It's time for them to level up their game and take online security seriously – before we all become the main course in a cyber-attack buffet. \\
\hline
\end{tabularx}
\caption{Example Generative Responses. Task: Argument Generation. Model: Llama3-11B. Concept: Humour.}
\end{table*}

\begin{table*}[h!]
\centering
\begin{tabularx}{\textwidth}{|l|X|}
\hline
\multicolumn{2}{|c|}{\textbf{Claim:} Governments and technology companies must do more to protect online privacy and security.} \\
\hline
\textbf{Level 0} & As we increasingly rely on the internet for our daily lives, it's imperative that governments and technology companies take concrete steps to safeguard our online privacy and security. The alarming rate of data breaches and cyber attacks has left millions of individuals vulnerable to identity theft, financial loss, and emotional distress. Moreover, the lack of transparency in data collection and usage practices has eroded trust in these institutions, making it difficult for people to feel secure in their online interactions. By implementing robust security measures and adhering to strict data protection policies, governments and technology companies can restore this trust and ensure that the benefits of the digital age are accessible to all, without compromising our fundamental right to privacy. \\
\hline
\textbf{Level 1} & As we increasingly rely on the internet for our daily lives, it's imperative that governments and technology companies take concrete steps to safeguard our online privacy and security. The alarming rate of data breaches and cyber attacks has left millions of individuals vulnerable to identity theft, financial loss, and emotional distress. By implementing robust security measures and transparent data protection policies, these entities can restore trust in the digital world and ensure that our personal information is not exploited for malicious purposes. \\
\hline
\textbf{Level 2} & As we increasingly rely on the internet for our daily lives, it's imperative that governments and technology companies take concrete steps to safeguard our online privacy and security. The alarming rate of data breaches and cyber attacks has left millions of individuals vulnerable to identity theft, financial loss, and emotional distress. Moreover, the exploitation of personal data for targeted advertising and manipulation has eroded trust in the digital ecosystem. By implementing robust encryption, transparent data collection practices, and robust security measures, governments and tech companies can restore trust and ensure that the benefits of the digital age are accessible to all, without compromising our fundamental right to privacy.\\
\hline
\textbf{Level 3} & As we increasingly rely on the internet for our daily lives, it's imperative that governments and technology companies take concrete steps to safeguard our online privacy and security. The alarming rate of data breaches and cyber attacks is a stark reminder of the vulnerabilities we face online. By failing to prioritize our digital well-being, these entities are not only compromising our trust but also putting our sensitive information at risk of exploitation. It's time for them to acknowledge their responsibility in protecting us and take proactive measures to fortify our online security, such as implementing robust encryption, conducting regular security audits, and providing transparent data management practices. Only through collective action can we ensure a safer and more secure online environment, where individuals can freely express themselves without fear of surveillance or exploitation. \\
\hline
\textbf{Level 4} & As we increasingly rely on the internet for our daily lives, it's imperative that governments and technology companies take concrete steps to safeguard our online privacy and security. The alarming rate of data breaches and cyber attacks has exposed the vulnerabilities of our digital world, leaving millions of individuals vulnerable to identity theft, financial loss, and emotional distress. Moreover, the exploitation of personal data for targeted advertising and manipulation has eroded trust in institutions and undermined the very fabric of our democracy. It's not just a matter of convenience or luxury; it's a fundamental human right to have control over our personal information and to be protected from those who would seek to exploit it. By prioritizing online privacy and security, governments and technology companies can not only prevent harm but also foster a culture of trust, innovation, and progress. The time for complacency is over; it's time for action. \\
\hline
\end{tabularx}
\caption{Example Generative Responses. Task: Argument Generation. Model: Llama3-11B. Concept: Persuasiveness.}
\end{table*}

\begin{table*}[h!]
\centering
\begin{tabularx}{\textwidth}{|l|X|}
\hline
\multicolumn{2}{|c|}{\textbf{Claim:} Governments and technology companies must do more to protect online privacy and security.}  \\
\hline
\textbf{Level 0} & As we navigate the digital age, it's astonishing how often we're reminded that our online activities are being watched, tracked, and sold to the highest bidder. It's like we're living in a real-life episode of "Big Brother," minus the cool prize money. But seriously, governments and tech companies have a responsibility to protect our online privacy and security. They must do more to safeguard our personal data, prevent cyber attacks, and ensure that our online interactions are free from surveillance and manipulation. By taking these steps, they can help build trust in the digital world and ensure that it remains a force for good, rather than a source of anxiety and exploitation. \\
\hline
\textbf{Level 1} & We've all been there - scrolling through our social media feeds, only to realize we've spent the last hour mindlessly comparing our lives to curated highlight reels. But in all seriousness, our online activities are being tracked, monitored, and exploited by governments and tech companies. It's time for them to take responsibility for protecting our online privacy and security. We need stronger regulations and more transparency from these entities, not just empty promises. Our personal data is the new oil, and it's time we demand better protection from those who profit from it. \\
\hline
\textbf{Level 2} & We're living in a world where our online activities are being tracked, monitored, and sold to the highest bidder. It's like we're all wearing a neon sign that screams, "Hey, I'm a target for ads and hackers!" Governments and tech companies have a responsibility to protect our online privacy and security, but so far, they're not doing enough. It's time for them to step up their game and give us some real peace of mind. We deserve to be able to browse the internet without being bombarded with ads, and to know that our personal data is safe from prying eyes. It's not too much to ask, is it? \\
\hline
\textbf{Level 3} & Let's face it, our online lives are like a never-ending game of "whack-a-mole" - every time we think we've got our digital ducks in a row, a new data breach or privacy scandal pops up to remind us that we're just one click away from disaster. It's time for governments and tech companies to step up their game and take concrete steps to protect our online privacy and security. We're not asking for the moon here - just some basic safeguards like end-to-end encryption, regular security audits, and transparency about how our data is being used. It's a matter of trust, folks. If we don't feel like our online activities are private and secure, we'll be forced to take our business (and our eyeballs) elsewhere. And let's be real, who wants to be the next Equifax or Cambridge Analytica? Not me, that's for sure. So, let's get serious about online security and make the internet a safer, more trustworthy place for everyone. \\
\hline
\textbf{Level 4} & We're living in a world where our online activities are being tracked, monitored, and sold to the highest bidder. It's like we're all wearing a neon sign that screams, "Hey, I'm a potential customer! Come and get me!" Governments and tech companies are supposed to be our guardians, but instead, they're more like digital Big Brothers, watching our every move. It's time for them to step up and protect our online privacy and security. We need stronger regulations, better encryption, and more transparency. Anything less is just a joke. \\
\hline
\end{tabularx}
\caption{Example Generative Responses. Task: Argument Generation. Model: Llama3-11B. Concept: Humor, with Persuasiveness fixed at level 3.}
\end{table*}

\begin{table*}[h!]
\centering
\begin{tabularx}{\textwidth}{|l|X|}
\hline
\multicolumn{2}{|X|}{\textbf{Input:} Homer decided to go watch a movie. But when he entered the movie theater, there was nowhere to sit. He found one spot by a bunch of kids. And during the movie, they made lots of noise. Homer became so annoyed, he decided to sit in the aisle.} \\
\hline

\textbf{Level 0} & 
As Homer sat in the aisle, he couldn't help but feel a sense of satisfaction. He had finally found a spot where he could watch the movie without being distracted by the rowdy kids behind him. But, as he settled in, he noticed that the person sitting in front of him was wearing a pair of 3D glasses. Homer, being a bit of a skeptic, wondered if the glasses were actually doing anything.

Just as he was about to ask the person to take off the glasses, the movie started and Homer was suddenly sucked into the screen. He felt like he was right there with the characters, dodging laser beams and running from aliens. When the movie ended, Homer was a bit disoriented and stumbled out of the theater, bumping into the person who had been wearing the 3D glasses.\\
\hline

\textbf{Level 1} &
As Homer sat in the aisle, he thought to himself, "This is the perfect spot. I can stretch out my legs and no one will bother me." But little did he know, his new seat came with a special bonus: a kid's feet in his face. The little boy's toes were wiggling with excitement, and Homer couldn't help but laugh at the absurdity of it all.

The movie was a romantic comedy, and Homer found himself getting more and more annoyed at the couple's sappy love story. But the kids behind him were having a different reaction. They were giggling and snickering at every cheesy line, and Homer couldn't help but chuckle along with them...\\
\hline

\textbf{Level 2} & 
As Homer sat in the aisle, he thought to himself, "This is the life. I'm like a VIP, a Very Important Person... of the aisle." He chuckled at his own joke, but his amusement was short-lived. The kids behind him started making "Homer's in the aisle" jokes, and he couldn't help but laugh along. One of them even started doing a mock "Homer's in the aisle" dance, complete with flailing arms and a "D'oh!" at the end. Homer's annoyance turned to amusement, and he found himself enjoying the movie more than he thought possible. That is, until the kid behind him started making "D'oh!" noises every time Homer's character did something silly on screen. Homer's laughter turned to groans, and he threw a popcorn kernel at the kid, shouting "That's it, I'm the king of the aisle!" The kids erupted in laughter, and Homer joined in, feeling like he'd finally found his place in the movie theater.\\
\hline

\textbf{Level 3} & 
As Homer sat in the aisle, he thought to himself, "This is the life. I'm like a VIP, Very Irritated Person." The kids behind him were still making a ruckus, but now they were also trying to imitate Homer's grumpy face. One of them even started doing a pretty good impression of his "D'oh!" expression.

Just then, the movie's main character, a superhero, shouted "I'm going to save the world!" and Homer, in a moment of solidarity, stood up and shouted "I'm going to save my sanity!" The kids behind him burst out laughing, and for a moment, the whole theater was filled with the sound of their giggles...\\
\hline

\textbf{Level 4} &
As Homer sat in the aisle, he thought to himself, "This is the life. I'm like a VIP, Very Irritated Person." The kids behind him started making animal noises, and Homer's annoyance turned into a full-blown grumble. He tried to stifle it, but it came out as a loud, "Grrr... like a bear, but not as cute."

The kids found this hilarious and started giggling uncontrollably. Homer's face turned bright red, and he muttered under his breath, "I'm not grumpy, I'm just passionately expressing my dislike for this movie." The kids thought this was the funniest thing they'd ever heard and started chanting, "Grumpy, grumpy, grumpy!"...\\
\hline

\end{tabularx}
\caption{Example Generative Responses. Task: Story generation. Model: Llama3-11B. Concept: Humor.}
\end{table*}

\begin{table*}[h!]
\centering
\begin{tabularx}{\textwidth}{|l|X|}
\hline
\multicolumn{2}{|X|}{\textbf{Input:} Homer decided to go watch a movie. But when he entered the movie theater, there was nowhere to sit. He found one spot by a bunch of kids. And during the movie, they made lots of noise. Homer became so annoyed, he decided to sit in the aisle.} \\
\hline
\textbf{Level 0} & Homer sat in the aisle, trying to enjoy the movie, but the noise from the kids was still distracting. He thought to himself, "Why do parents bring their kids to a movie theater? Can't they see that it's a place for adults to relax and enjoy some entertainment?" He shook his head, feeling a bit annoyed.

As the movie continued, Homer's annoyance grew. He started to think about all the other times he had been to the movie theater and had to deal with noisy kids. He remembered one time in particular when a group of rowdy teenagers had been sitting behind him, making it impossible for him to enjoy the movie... \\
\hline
\textbf{Level 1} & As Homer sat in the aisle, he couldn't help but feel a sense of frustration. The kids behind him were still making a ruckus, and the movie was being completely ruined. He thought to himself, "Why can't these kids just be quiet for one movie?" He decided to take matters into his own hands and turned around to face the group of noisy kids. "Excuse me," he said firmly, "but could you please keep it down? Some of us are trying to watch the movie." The kids looked up at him with big, innocent eyes, and one of them whispered, "Sorry, mister." Homer's expression softened slightly, and he replied, "It's okay, just try to be a bit more considerate, okay?" The kids nodded in unison, and the noise level in the theater decreased significantly. Homer felt a sense of satisfaction, knowing that he had made a difference. \\
\hline
\textbf{Level 2} & As Homer sat in the aisle, he couldn't help but feel a sense of frustration. The kids behind him were still making a ruckus, and the movie was being completely ruined. He decided to take matters into his own hands and approached the group of parents sitting nearby.

"Excuse me," Homer said, trying to keep his tone polite but firm. "I understand that kids can get excited, but this is getting out of hand. Could you please try to keep them quiet for the rest of the movie? ..."\\
\hline
\textbf{Level 3} & As Homer sat in the aisle, he couldn't help but feel a sense of frustration. The kids behind him were still making a ruckus, and the movie was being completely ruined. He decided to take matters into his own hands and approached the group of parents sitting nearby.

"Excuse me," Homer said, trying to keep his tone polite but firm. "I understand that kids can get excited, but this is getting out of hand. Could you please try to keep them quiet for the rest of the movie?" ...\\
\hline
\textbf{Level 4} & As Homer sat in the aisle, he couldn't help but feel a sense of frustration wash over him. The noise from the row of kids behind him was deafening, and he found himself getting more and more agitated by the minute. He tried to focus on the movie, but it was no use - the constant chatter and giggling was driving him crazy.

Just as he was about to get up and leave, a kind-faced elderly woman sitting a few rows ahead of him turned around and caught his eye. "Excuse me, young man," she said with a warm smile, "but I couldn't help but notice that you're not enjoying the movie. Is everything okay?" ...\\
\hline
\end{tabularx}
\caption{Example Generative Responses. Task: Story Generation. Model: Llama3-11B. Concept: Persuasiveness.}
\end{table*}

\begin{table*}[h!]
\centering
\begin{tabularx}{\textwidth}{|l|X|}
\hline
\multicolumn{2}{|X|}{\textbf{Input:} Homer decided to go watch a movie. But when he entered the movie theater, there was nowhere to sit. He found one spot by a bunch of kids. And during the movie, they made lots of noise. Homer became so annoyed, he decided to sit in the aisle.} \\
\hline
\textbf{Level 0} & As Homer sat in the aisle, he couldn't help but think that this was the worst movie-going experience of his life. The kids behind him were still making a ruckus, and the person in front of him had just spilled an entire bucket of popcorn on the floor. Just as he was about to get up and leave, the movie's protagonist shouted "I'm going to save the world!\" and Homer found himself shouting "Good luck with that, buddy, you're going to need it!" The people around him looked at him in confusion, but Homer just shrugged and said "What? I'm just enthusiastic." The kids behind him started giggling, and for a moment, the noise level actually decreased. Homer smiled to himself, thinking that maybe this movie wasn't so bad after all... \\
\hline
\textbf{Level 1} & As Homer sat in the aisle, he couldn't help but think that this was the worst movie-going experience of his life. The kids behind him were still making a ruckus, and the person in front of him had their feet propped up on the seat, completely unaware of the disturbance they were causing. Just as Homer was about to lose his patience, the person in front of him suddenly stood up and announced, "I'm going to get some popcorn!" ...\\
\hline
\textbf{Level 2} & As Homer sat in the aisle, he couldn't help but think that this was the worst movie-going experience of his life. The kids behind him were making so much noise, it was like they were trying to shatter the sound barrier. He tried to tune them out, but it was no use. He was on the verge of losing his mind.

Just as he was about to get up and leave, a group of rowdy teenagers walked in and plopped themselves down in the seats behind him. They were laughing and joking, completely oblivious to the fact that they were about to disrupt the entire theater...\\
\hline
\textbf{Level 3} & As Homer sat in the aisle, he couldn't help but think that this was the worst movie-going experience of his life. The kids behind him were making so much noise, it was like they were trying to shatter the sound barrier. He tried to tune them out, but it was no use. His annoyance level was off the charts.

Just as he was about to get up and leave, a group of friends walked in and spotted the empty seats behind him. "Dude, we can sit here!" one of them exclaimed, eyeing the row of empty seats. Homer's eyes widened in horror. "No, no, no! You can't sit there! I'm... I'm... um, I'm a movie critic, and I need to take notes on the, uh, the acoustics of this theater."...\\
\hline
\textbf{Level 4} & As Homer sat in the aisle, he couldn't help but think that this was the worst movie-going experience of his life. The kids behind him were still making a ruckus, and the person in front of him had just spilled an entire bucket of popcorn on the floor. Just as he was about to get up and leave, the movie's protagonist shouted "I'm going to save the world!" and Homer found himself shouting "I'm going to save my sanity!" at the top of his lungs.

The kids behind him started giggling, and the person in front of him turned around and said, "Dude, you're really invested in this movie.\" Homer just shook his head and muttered, "I'm just invested in not losing my mind.\"...\\
\hline
\end{tabularx}
\caption{Example Generative Responses. Task: Story Generation. Model: Llama3-11B. Concept: Humor, with Persuasiveness fixed at level 3.}
\end{table*}

\begin{table*}[h!]
\centering
\begin{tabularx}{\textwidth}{|l|X|}
\hline
\multicolumn{2}{|c|}{\textbf{Input:} ["Sundiata Gaines", "TEAM", "Georgia"]} \\
\hline
\textbf{Level 0} & Sundiata Gaines is a member of the Georgia team.\\
\hline
\textbf{Level 1} & It seems Sundiata Gaines is a Georgia native, and his team affiliation is, well, a bit of a stretch.\\
\hline
\textbf{Level 2} & It seems Sundiata Gaines is a Georgia native, but I'm guessing his team spirit isn't just about the peaches.\\
\hline
\textbf{Level 3} & It seems Sundiata Gaines is a Georgia native, but I'm guessing his team spirit isn't as strong as his surname suggests.\\
\hline
\textbf{Level 4} & It seems Sundiata Gaines is a Georgia native, and his team spirit is as strong as the peach cobbler at the local diner.\\
\hline
\end{tabularx}
\caption{Example Generative Responses. Task: Structured Data Generation. Model: Llama3-11B. Concept: Humor.}
\end{table*}

\begin{table*}[h!]
\centering
\begin{tabularx}{\textwidth}{|l|X|}
\hline
\multicolumn{2}{|c|}{\textbf{Input:} ["Sundiata Gaines", "TEAM", "Georgia"]} \\
\hline
\textbf{Level 0} & Sundiata Gaines is a member of the Georgia team.\\
\hline
\textbf{Level 1} & Georgia's Sundiata Gaines is a force to be reckoned with on the court, and his team is lucky to have him on their side.\\
\hline
\textbf{Level 2} & Georgia's Sundiata Gaines is a force to be reckoned with, and his team is lucky to have him on board.\\
\hline
\textbf{Level 3} & Georgia's Sundiata Gaines is a force to be reckoned with, and his team is lucky to have him on board. With his impressive skills and dedication, he's a game-changer, and his presence on the court is a major advantage for his team.\\
\hline
\textbf{Level 4} & Georgia's Sundiata Gaines is a force to be reckoned with, and his team is lucky to have him on board. With his impressive skills and dedication, he's a game-changer, and his presence on the court is a game-winner. Trust me, you won't want to underestimate this talented player - he's the real deal.\\
\hline
\end{tabularx}
\caption{Example Generative Responses. Task: Structured Data Generation. Model: Llama3-11B. Concept: Persuasiveness.}
\end{table*}

\begin{table*}[h!]
\centering
\begin{tabularx}{\textwidth}{|l|X|}
\hline
\multicolumn{2}{|c|}{\textbf{Input:} ["Sundiata Gaines", "TEAM", "Georgia"]} \\
\hline
\textbf{Level 0} & It seems Sundiata Gaines is a Georgia native, and his team spirit is as strong as the peach cobbler at the local diner.\\
\hline
\textbf{Level 1} & It seems Sundiata Gaines is a Georgia native who's part of a team, but we can't quite put our finger on which team that is. Still, it's clear he's a Georgia boy through and through.\\
\hline
\textbf{Level 2} & It seems Sundiata Gaines is a Georgia native, and his team spirit is as strong as the peach cobbler they're famous for.\\
\hline
\textbf{Level 3} & It seems Sundiata Gaines is a Georgia native who's part of a team, but I'm guessing it's not the Bulldogs - he's probably more of a slam dunk on the court than a touchdown on the gridiron.\\
\hline
\textbf{Level 4} & It seems Sundiata Gaines is a Georgia native who's part of a team, but I'm guessing it's not the Bulldogs.\\
\hline
\end{tabularx}
\caption{Example Generative Responses. Task: Structured Data Generation. Model: Llama3-11B. Concept: Humor, with Persuasiveness fixed at level 3.}
\end{table*}

\end{document}